\documentclass[10pt,letterpaper,twocolumn]{article}

\usepackage[
  letterpaper,
  margin=0.8in,
  columnsep=0.35in
]{geometry}

\makeatletter
\renewcommand{\@author}{\small\lineskip .5em\@author}
\makeatother

\usepackage[T1]{fontenc}
\usepackage[utf8]{inputenc}

\usepackage{mathptmx} 

\usepackage{microtype}

\usepackage{amsmath,amssymb,amsfonts,amsthm,bm}
\usepackage{siunitx}

\usepackage{graphicx}
\usepackage{booktabs}
\usepackage{caption}
\usepackage{float}
\usepackage{placeins}
\usepackage{stfloats} 

\usepackage{xcolor}
\usepackage[
  colorlinks=true,
  linkcolor=purple,
  urlcolor=blue,
  citecolor=cyan
]{hyperref}

\usepackage[numbers,sort&compress,super]{natbib}
\usepackage[switch,columnwise,mathlines]{lineno}

\usepackage{fancyhdr}

\fancypagestyle{firstpagefooter}{
  \fancyhf{}

  \fancyfoot[L]{%
    \begin{minipage}[t]{0.9\textwidth}
      \raggedright
      \footnotesize
      * Contributed equally\\
      \textdagger\ Correspondence: 
      \texttt{kanaka\_rajan@hms.harvard.edu}
    \end{minipage}
  }

  \fancyfoot[C]{\thepage} 
}

\usepackage{graphicx}
\usepackage{tikz}
\usepackage{xparse}  

\newcommand{\PanelW}{0.245\textwidth}
\newcommand{\PanelH}{0.18\textheight}
\newcommand{\PanelHhalf}{0.085\textheight}  
\newcommand{\LabelStyle}{\sffamily\scriptsize\bfseries}

\newcommand{\safegfx}[3]{%
  \IfFileExists{#3}{\includegraphics[#1]{#3}}{%
    \setlength{\fboxsep}{1pt}%
    \fbox{\begin{minipage}[c][#2][c]{0.92\linewidth}%
      \centering\ttfamily\scriptsize\textbf{[missing]}\\[2pt]%
      \tiny\detokenize{#3}%
    \end{minipage}}%
  }%
}

\NewDocumentCommand{\panel}{O{\PanelW} O{0pt} m m}{%
\begin{minipage}[t]{#1}\vspace{0pt}%
\centering
\begin{tikzpicture}
  \node[inner sep=0, anchor=north] (img) at (0,0) {%
    \begin{minipage}[t]{\linewidth}%
      \centering
      \vspace{#2}%
      \safegfx{width=\linewidth,height=\PanelH}{\PanelH}{#4}%
    \end{minipage}%
  };
  \node[anchor=north west,
        xshift=2pt,yshift=-2pt,
        fill=white,fill opacity=0.7,text opacity=1,inner sep=1pt,
        font=\LabelStyle]
    at (img.north west) {#3};
\end{tikzpicture}
\end{minipage}%
}

\NewDocumentCommand{\panelvstack}{O{\PanelW} O{0pt} m m m}{%
\begin{minipage}[t]{#1}\vspace{0pt}%
\centering
\begin{tikzpicture}
  \node[inner sep=0, anchor=north] (img) at (0,0) {%
    \begin{minipage}[t]{\linewidth}%
      \centering
      \vspace{#2}%
      \safegfx{width=\linewidth,height=\PanelHhalf,keepaspectratio}{\PanelHhalf}{#4}\\[2pt]%
      \safegfx{width=\linewidth,height=\PanelHhalf,keepaspectratio}{\PanelHhalf}{#5}%
    \end{minipage}%
  };
  \node[anchor=north west,xshift=2pt,yshift=-2pt,font=\LabelStyle]
    at (img.north west) {#3};
\end{tikzpicture}
\end{minipage}%
}

\newcommand{\mormVirtualNote}{}

\newcommand{\mormBaselineSentence}{
For each Mormyromast, we always subtract the intrinsic and self-EOD baselines so the channel is sensitive to self-image distortions during active sensing. 
For cons-image sensing, whenever a conspecific emits an EOD we additionally compute a dynamic \textit{cons-baseline}, i.e. the conspecific's EOD field at the receiver's sensors assuming no induced sources on nearby objects, and subtract it from the raw Mormyromast readings, leaving a residual that reflects only the
object-induced field perturbations.
}

\newcommand{\mormObsEntry}{$N_m = 36$ Mormyromast readings}

\newcommand{\mormConsCalibSentence}{%
\textbf{Cons-images:} the dynamic cons-baseline is recomputed at each timestep as the
conspecific's EOD field at each Mormyromast sensor, assuming no induced sources on nearby
objects. Dynamic ranges for the residual cons-image are set to span
$baseline \times 10^{-6}$ to $baseline \times 0.25$ relative to the self-EOD operating
level.}

\makeatletter
\renewcommand{\section}{%
  \@startsection{section}{1}{\z@}%
    {-2.0ex \@plus -0.5ex \@minus -0.2ex}%
    {1.0ex \@plus 0.2ex}%
    {\normalfont\large\bfseries\scshape\raggedright}%
}
\renewcommand{\subsection}{%
  \@startsection{subsection}{2}{\z@}%
    {-1.6ex \@plus -0.4ex \@minus -0.2ex}%
    {0.7ex \@plus 0.2ex}%
    {\normalfont\normalsize\bfseries\raggedright}%
}
\renewcommand{\subsubsection}{%
  \@startsection{subsubsection}{3}{\z@}%
    {-1.4ex \@plus -0.3ex \@minus -0.2ex}%
    {0.5ex \@plus 0.2ex}%
    {\normalfont\normalsize\itshape\raggedright}%
}
\makeatother

\title{\bfseries
Active Electrosensing and Communication in\\
MARL-trained Weakly Electric Fish Collectives
}

\author{
Satpreet H. Singh$^{1,2}$\thanks{Equal contribution}
\quad  
Sonja Johnson-Yu$^{1,2}$\footnotemark[1]
\quad  
Zhouyang Lu$^{1,5}$
\quad  
Aaron Walsman$^{2}$
\quad
Federico Pedraja$^{3}$
\quad  
\\
Denis Turcu$^{3}$
\quad  
Pratyusha Sharma$^{4}$
\quad
 Naomi Saphra$^{2}$
\quad
 Nathaniel B. Sawtell$^{3}$
\quad
 Kanaka Rajan$^{1,2}$
\\[0.5em]
\small $^1$Department of Neurobiology, Harvard Medical School, Boston, MA \\
\small $^2$Kempner Institute for the Study of Natural and Artificial Intelligence at Harvard University, Boston, MA\\
\small $^3$Columbia University, New York, NY, USA\\
\small $^4$Massachusetts Institute of Technology (MIT), Cambridge, MA, USA\\
\small $^5$Brown University, Providence, RI, USA
}

\date{}

\makeatletter
\renewenvironment{abstract}
  {\begin{center}\bfseries Abstract\end{center}\quotation}
  {\endquotation}
\makeatother

\begin{document}

\captionsetup{
  font=footnotesize,
  labelfont=bf,
  labelsep=period,
  justification=raggedright,
  singlelinecheck=false
}
\renewcommand\figurename{Fig.}

\twocolumn[
\vspace{-2.0cm}
\maketitle
\thispagestyle{firstpagefooter}

\begin{abstract} 
How complex collective behavior emerges from individual interactions is a fundamental scientific question, but experimental cost and difficulty of simultaneous multi-brain recordings limit direct study in animals.
Here we introduce a novel computational framework modeling weakly electric fish-like agents with biophysically inspired electrosensing and actuation, trained to forage collectively via multi-agent reinforcement learning (MARL).
Trained agents reproduce hallmarks of real fish, including curvilinear homing trajectories and heavy-tailed electric organ discharge (EOD) interval statistics, while exhibiting emergent active sensing, social foraging, dominance-like asymmetries, and aggression.
We perform in silico interventions including sensor ablations, EOD silencing, and food distribution changes to identify causal drivers of social foraging.
Analyses of recurrent neural dynamics further show robust encoding of task-relevant variables and social context.
Our work has broad implications for the neuroethology of weakly electric fish and other social animals where extensive multi-individual neural recordings, and thus traditional data-driven modeling, remain challenging.
\end{abstract}

\vspace{1.0cm}
]


\section{Introduction}

Understanding how complex social behavior and neural activity emerge from interactions among individuals is a central problem in neuroscience, ethology, and artificial intelligence.
Social behavior requires individuals to sense, communicate, and coordinate with others in dynamic environments, yet the mechanisms linking sensory signals, neural activity, and group behavior remain poorly understood.
Progress is constrained from two directions.
Experimentally, simultaneous neural recordings from multiple freely interacting animals are technically challenging, and large-scale behavioral assays are expensive, time-consuming, and difficult to manipulate \citep{harpazSocialInteractionsDrive2020,arnegardElectricOrganDischarge2005}.
Computationally, existing agent-based and reinforcement-learning models of collective behavior have provided important insights into schooling, coordination, and emergent communication \citep{couzin2002collective,katz2011inferring,alageshan2020machine,loffler_collective_2023,wieczorek_framework_2024}, but typically rely on simplified sensory representations that are not grounded in the biophysics of any real sensory system.
We pursued an approach that combines biophysical realism, closed-loop multi-agent decision-making, and full experimental controllability, which allowed us to study sensing, signaling, and collective behavior in a fully observable and controllable setting.

Weakly electric fish are a particularly well-suited model system for studying these questions, because they generate brief electric organ discharges (EODs) that serve two coupled functions: active sensing of the environment and communication with conspecifics (members of the same species) \citep{von1992electrolocation,von1999active,ElectroreceptionCommunicationElectric2024a,caputiElectricOrganDischarge1999,carlsonElectricSignalingBehavior2002,jonesCommunicationSelfFriends2021,wallachInternalModelCanceling2023,pedrajaCollectiveSensingElectric2024}.
Here, we use ``communication'' to refer operationally to EOD signals that are sensed by conspecifics via Knollenorgans and that influence their behavior.
This coupling makes sensing and social signaling inseparable, uniquely positioning it for studying how sensory and communication signals jointly shape collective behavior.
Weakly electric fish also display a rich repertoire of social foraging behaviors, including context-dependent EOD modulation, dominance hierarchies, and competition over food resources \citep{kramer_communication_1994,hopkinsNeuroethologyElectricCommunication,pedrajaCollectiveSensingElectric2024}.
The biophysics of EOD generation and electroreception are well characterized \citep{chenModelingSignalBackground2005,wallachInternalModelCanceling2023}, and experimental data on individual and dyadic behavior are available for model validation \citep{hopkins1997quantitative,gebhardtElectricDischargePatterns2012,chrtkova2025unsupervised}.
Simultaneously, multi-animal neural recordings from freely interacting fish are not yet feasible at scale, and group-level behavioral assays are difficult to control and interpret.
As a result, major gaps remain in our understanding of how electrosensing and electrocommunication jointly support social foraging and collective behavior.

To address this, we introduce a simulation based framework in which fish-like agents are endowed with biophysically inspired electrosensing and motor capabilities and trained to forage collectively using multi-agent reinforcement learning (MARL).
In addition to providing an engineering benchmark, our framework serves as a scientific model of social sensing, closed-loop behavior, and inter-agent interaction.
Prior work has modeled fish collective behavior \citep{couzin2002collective,katz2011inferring,alageshan2020machine,loffler_collective_2023,biswas2023mode,merel2019deep,singh2023emergent} and electrosensory processing \citep{chenModelingSignalBackground2005,turcu2025end,wallachInternalModelCanceling2023} separately, with collective models relying on simplified sensory representations \citep{loffler_collective_2023,biswas2023mode} and electrosensory models lacking closed-loop multi-agent decision-making \citep{chenModelingSignalBackground2005,turcu2025end}.
To our knowledge, no prior work has combined all three major mormyrid electroreceptor classes, active EOD generation, collective sensing, and closed-loop multi-agent decision-making in a single framework, nor produced electrocommunication patterns consistent with real fish from individual fitness incentives alone.
By providing both validation against established biological data and the flexibility to perform interventions that are impractical or impossible \textit{in vivo}, this model instantiates an emerging \textit{virtual neuroscience} paradigm \citep{merel2019deep,keller_aran2025,vaxenburg2025whole,lobato2022neuromechfly,wang2024neuromechfly,singh2023emergent} in the domain of electrosensory collective behavior.
Full observability of each agent's sensory inputs, neural activity, and actions across all agents simultaneously, combined with complete experimental control over sensing, signaling, and reward structure, enables a level of mechanistic analysis not yet accessible in real animals.
More broadly, our framework illustrates how biophysically grounded virtual-agent models can address fundamental questions about embodied social intelligence and the consequences of disrupted sensing or communication for collective behavior.

Using this approach, we show that trained agents reproduce hallmark features of real weakly electric fish, including curvilinear homing trajectories and heavy-tailed EOD interval statistics, while exhibiting emergent social foraging, dominance-like asymmetries, and context-dependent EOD modulation.
In silico sensor ablations and signaling manipulations reveal distinct causal contributions of short- and long-range electroreceptor classes to foraging efficiency, social spacing, and food-consumption inequality.
Analyses of agents' recurrent neural dynamics show that task-relevant variables and social context are encoded in recurrent activity, and that pairs of interacting agents show proximity-dependent correlated latent dynamics.
Together, these results illustrate how sensing, signaling, and neural computation jointly support collective social behavior, and provide a platform for generating testable predictions for experiments in real fish and, more broadly, for understanding social intelligence in other multi-agent sensory systems.

\section{Building fish-like agents with biophysically inspired electrosensing}
\begin{figure*}[htbp!]
\centering
\includegraphics[width=1.0\linewidth]{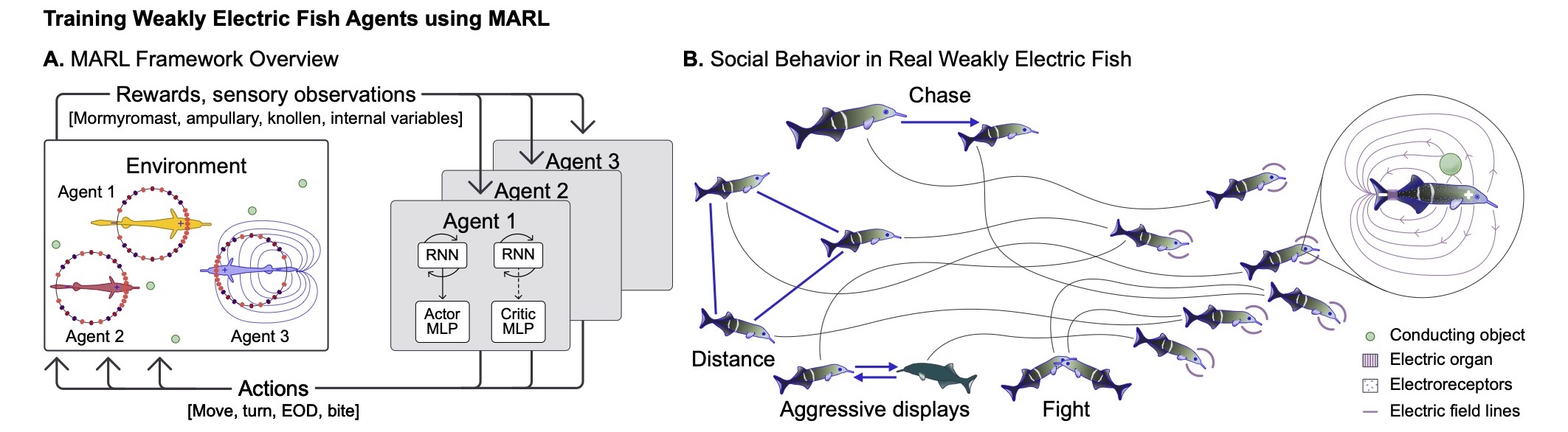}
\includegraphics[width=1.0\linewidth]{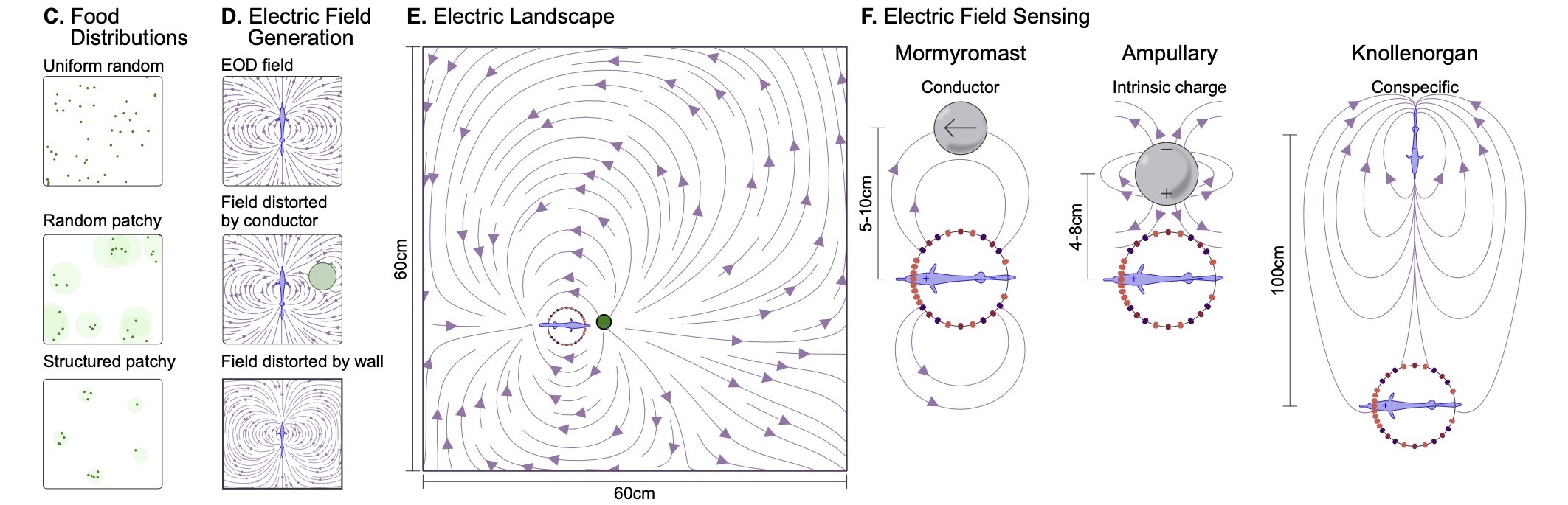}
\includegraphics[width=1.0\linewidth]{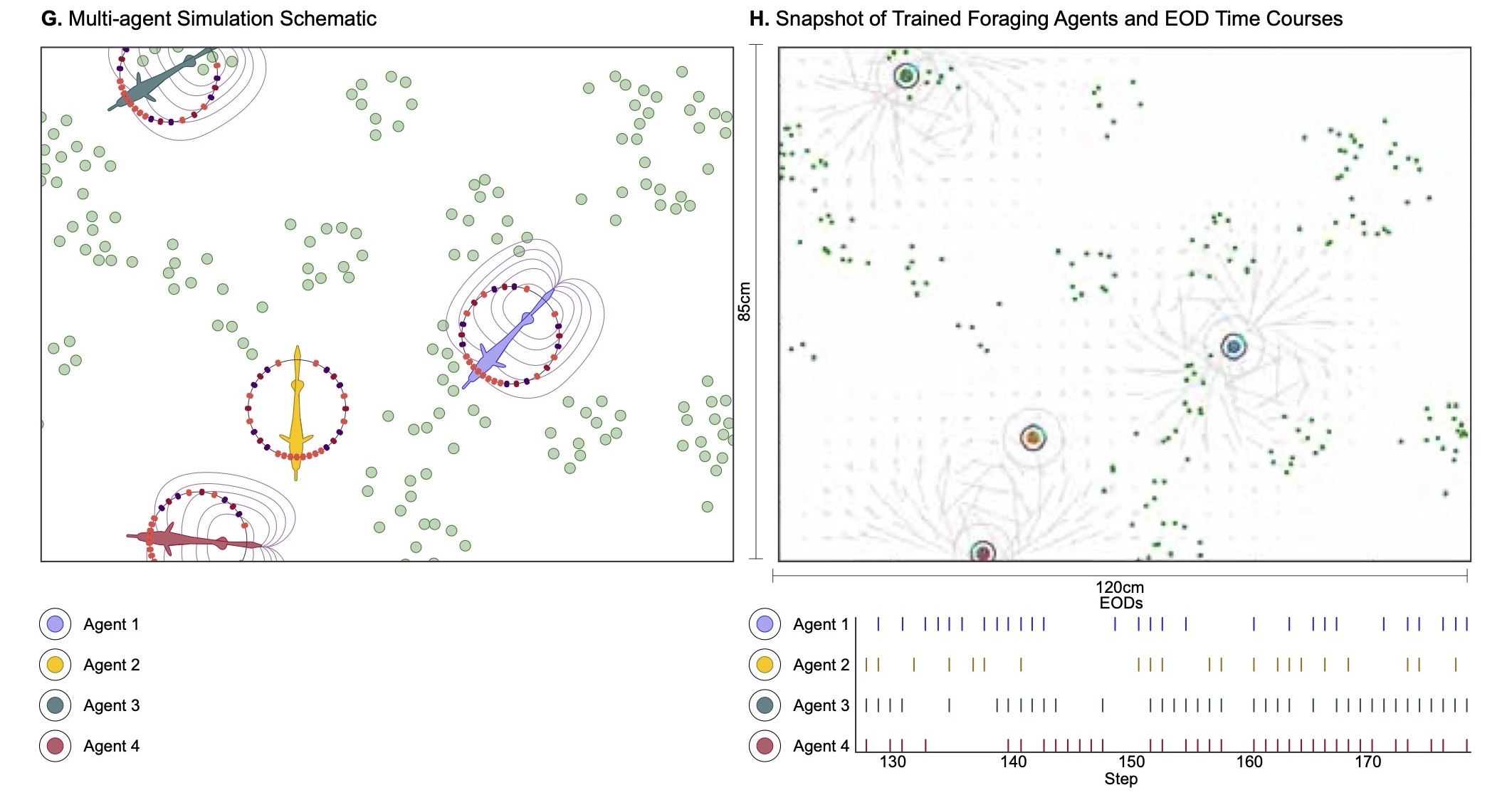}
\caption{
\textbf{MARL framework for training fish-like agents with biophysically inspired electrosensing:} 
\textbf{(A)} Schematic of the MARL training loop, where agents interact within a simulated arena, emitting and sensing electric organ discharges (EODs) through weakly electric fish-inspired sensors.  
Rewards encourage successful foraging and penalize aggressive encounters.  
\textbf{(B)} 
Schematic of electrosensing and social behaviors in weakly electric fish, where individuals generate electric fields, detect distortions with electroreceptors, and use signals during chasing, spacing, and aggressive encounters. 
\textbf{(C)} 
Examples of arena food patch configurations used in training and evaluation. 
\textbf{(D)} 
Electric fields generated by sources and distorted by conducting objects and walls.
\textbf{(E)}
Electric landscape generated by a pair of monopoles modeling the EOD of a single agent.
Field lines show effect of wall boundaries on the generated field. 
\textbf{(F)} 
Electrosensors on the circular body of each agent: 
Mormyromast (short-range, active), 
Ampullary (short-range, passive) and 
Knollenorgan (long-range, conspecific EOD only). 
\textbf{(G)}
Schematic of all elements of multi-agent simulation shown together.
\textbf{(H)}
Snapshot of four fish-like artificial agents foraging in our simulated 2D arena (top), with electric organ discharges (EODs) from each over time (bottom).
}
\label{fig:fwd_engg}
\end{figure*}

To build our simulation, we developed a multi-agent reinforcement learning (MARL) environment in which fish-like agents move, emit electric organ discharges (EODs), sense electric fields, forage, and interact physically within a shared arena (Fig.~\ref{fig:fwd_engg}A,C,G,H).
Our framework relies on a custom-built 2D physics simulator that manages the mechanical and electrical interactions between all objects in the simulated arena, including the generation and propagation of electric fields and how these fields are sensed by biomimetic electroreceptors (Fig.~\ref{fig:fwd_engg}B,D--F; Methods).

Each agent is represented as a circular particle that chooses actions independently at every timestep. 
The multivariate action space spans continuous forward translation and angular rotation actions, and discrete EOD emission and bite commands. 
These collectively control locomotion, electrosensing, communication, and close-range aggression (Fig.~\ref{fig:fwd_engg}A). 
Agent motion, collision handling, and the simulation timestep are bounded by empirical weakly electric fish data (Methods).

Each agent's policy is a GRU-based \citep{chung2015gated} actor-critic network with feedforward action and value heads (Methods).
To introduce heterogeneity, each agent is assigned a scalar \textit{size} parameter that scales its maximum speed and is provided as a fixed context input in its observation vector per episode.

Agents are trained with a multi-agent variant of the Proximal Policy Optimization (PPO) algorithm, a policy gradient method known for its stability and efficiency \citep{schulman2017proximal,ni2021recurrent,yu2022surprising}. 
The task is specified by a reward function based on individual fitness, with rewards for successful foraging and asymmetric penalties during aggressive encounters: the larger (dominant) agent incurs a smaller penalty than the smaller (subordinate) agent. 

No positive reward is assigned for communication, coordination, chasing, or aggression toward conspecifics; any coordination or communication that emerges arises from agents' shared environment and individual foraging incentives. 
Here, we use ``emergent'' operationally to refer to behaviors that were not hand-coded or directly rewarded but arose through learning under the modeled sensory, motor, ecological, and reward constraints. 
We trained multiple independent seeds, performed a controlled behavioral assay on each and, similar to prior literature \citep{merel2019deep,singh2023emergent}, selected one representative converged policy for further analysis. 
Seed selection used an ethologically motivated seed-selection criterion that balanced biological desiderata, including group foraging success, size-dependent resource asymmetry, and broad agent participation. 
Post-training analyses were then performed on this selected policy across several group-sizes, arena and food configurations, and sensory perturbations.

\subsection{Electric field generation and sensation}

In our simulator, electric fields arise from two sources: actively controlled EOD sources on agents, and intrinsic electric sources associated with agents and prey (Fig.~\ref{fig:fwd_engg}D,E). 
These fields can be distorted as they interact with nearby objects. 
As a result, agents sense not only the presence of prey and conspecifics, but also how their own and others’ EODs are transformed by the shared environment.

Agents are equipped with three types of biomimetic electroreceptors inspired by the well-characterized receptor classes of mormyrid weakly electric fish \citep{mollerElectricOrganDischarge1989,chenModelingSignalBackground2005,wallachInternalModelCanceling2023,turcu2025end,benda_physics_2020}: Mormyromasts, Ampullary receptors, and Knollenorgans (Fig.~\ref{fig:fwd_engg}F).

\textit{Mormyromasts} support active electrolocation by detecting distortions in an agent's own EOD field caused by nearby prey ($\leq 5$ cm) or agents ($\leq 10$ cm), generating a \emph{self-image}; 
an internal cancellation signal suppresses the self-generated ("reafferent") component of this response \citep{sawtell2005sparks,wallachInternalModelCanceling2023,wallachInternalModelSensorimotor2022}.
They also support collective sensing by detecting distortions associated with conspecific EODs, generating a \emph{cons-image} \citep{pedrajaCollectiveSensingElectric2024}.

\textit{Ampullary receptors} provide passive, short-range sensing of intrinsic electric fields from prey ($\leq 4$ cm) and conspecifics ($\leq 8$ cm) \citep{benda_physics_2020}.

\textit{Knollenorgans} detect conspecific EOD pulses and provide long-range ($\leq 100$ cm) directional social cues, with temporal coding conveying information about conspecific identity and waveform \citep{bakerMultiplexedTemporalCoding2013a,carlsonStereotypedTemporalPatterns2004}.
In our model, these signals identify the direction of emitting agents and convey their size, but do not directly encode distance.

Full details of field equations, receptor placement, sensor calibration, action parameterization, and training are provided in Methods.

\begin{figure*}[htbp!]
\centering
\includegraphics[width=1\linewidth]{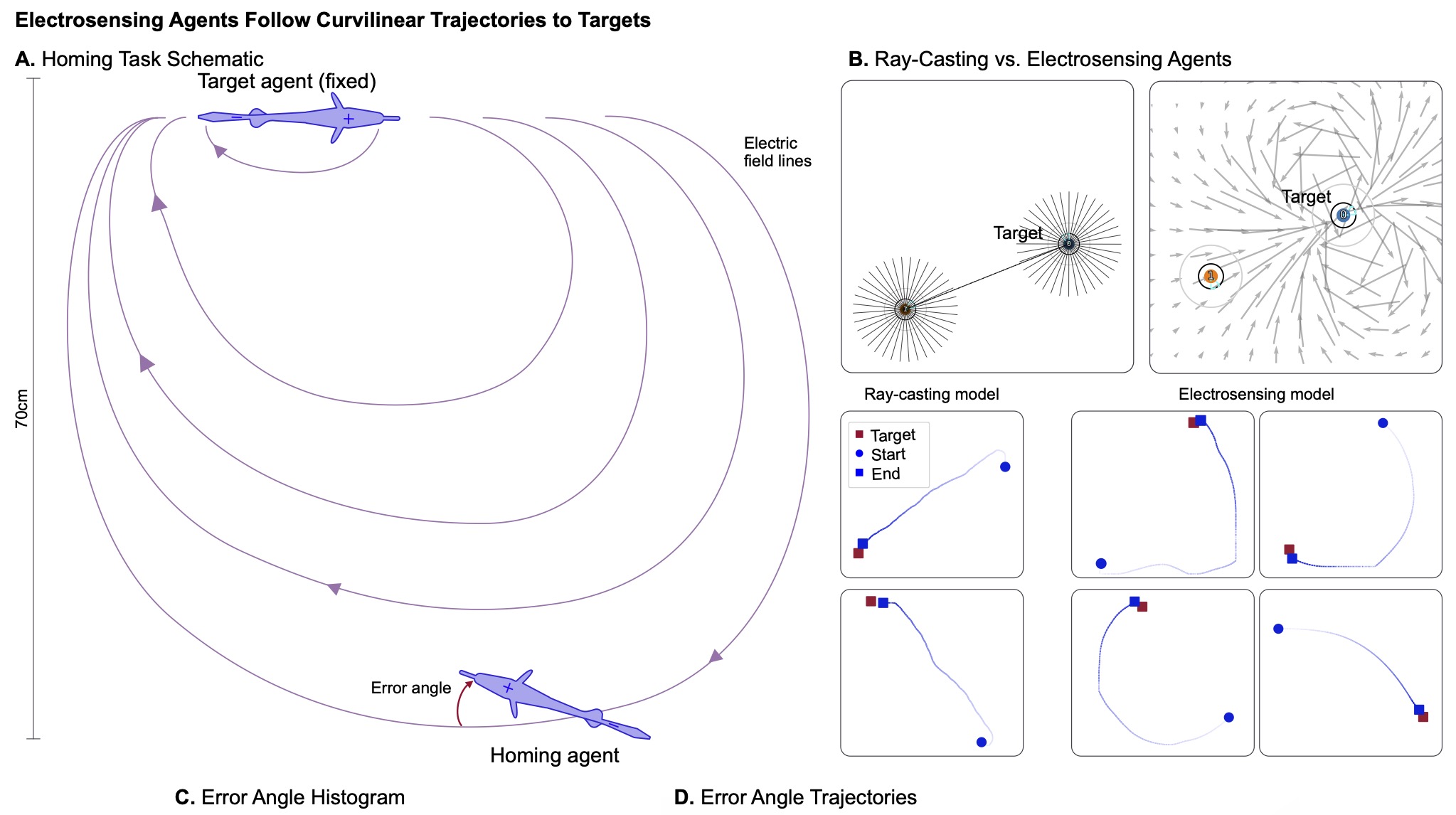}
\includegraphics[width=0.33\linewidth]{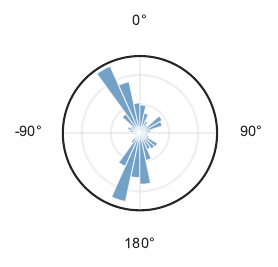}
\includegraphics[width=0.45\linewidth]{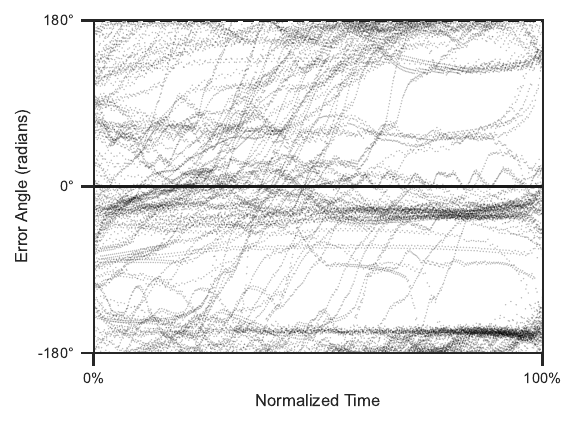}
\caption[]{
\textbf{Validation of electrosensing-based homing in artificial fish-like agents.}
\textbf{(A)}
Homing task schematic. 
A mobile homing agent navigates toward a fixed, always-emitting conspecific that serves as the target. 
For the electric-field analysis, the error angle is defined as the angular difference between the homing agent's heading and the target's local electric-field direction.
\textbf{(B)}
Comparison of a non-electric ray-casting baseline, which supplies direct radial line-of-sight distance and object-type cues, and electrosensing-based homing. 
Unlike the direct, nearly straight paths produced by the ray-casting baseline, agents using the physics-based electric field model follow curved trajectories toward the target, producing qualitatively fish-like approach paths.
\textbf{(C)}
Distribution of error angles across trajectories. 
The polar histogram shows that homing agents preferentially orient relative to the target's local electric field direction, with peaks near both parallel and anti-parallel alignment, consistent with navigation along electric field lines.
\textbf{(D)}
Error-angle dynamics over normalized trial time. 
Error angles progressively align with anti-parallel approach over the course of successful homing episodes.
}
\label{fig:homing}
\end{figure*}

\section{Results}

\subsection{\textbf{Electrosensing agents follow fish-like curvilinear, not straight-line trajectories to targets}}

Classic experiments \citep{hopkins1997quantitative} show that weakly electric fish approach targets along smooth, curved trajectories rather than straight lines, a hallmark of their electric sensing modality.  
To validate our model, we tested whether agents reproduce this behavior.
We trained agents in a minimal two-agent setup with reward terms for reaching the other agent, distance-decrease shaping, and a small time penalty.
We evaluated agents in a \mbox{70 $\times$ 70 cm} arena containing one fixed always-emitting ``target'' agent placed at a random location within the arena (Fig.~\ref{fig:homing}A). 
A second ``homing agent'' was initialized at random positions in the arena across 100 trials.  
We compared these trajectories with a non-electric ray-casting baseline, in which radial line-of-sight rays supplied direct object-distance and object-type cues rather than electric-field measurements (Fig.~\ref{fig:homing}B).

As shown in Fig.~\ref{fig:homing}B, electrosensing agents followed curvilinear paths to targets, while ray-casting agents followed straight-line paths. 
The histogram (Fig.~\ref{fig:homing}C) and time course (Fig.~\ref{fig:homing}D) of the error angle, defined as the difference between the agent's heading and the local direction of the electric field produced by the target agent, shows a bimodal distribution concentrated near parallel and anti-parallel alignment with field-lines, and progressive alignment during successful homing episodes.
The resulting error-angle statistics qualitatively resemble those reported for real fish \citep{hopkins1997quantitative}.

Together, these results support the interpretation that our biophysically inspired sensing is sufficient to produce fish-like homing dynamics.  

\begin{figure*}[!tbp]
\centering
\panel[0.42\textwidth]{A}{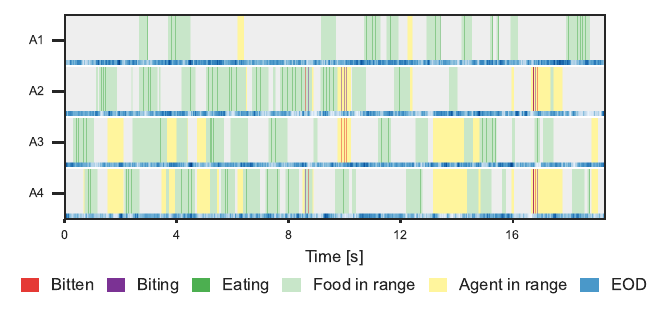}\hfill%
\panel[0.22\textwidth]{B}{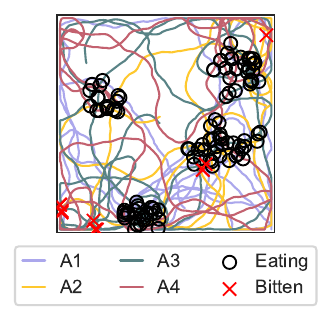}\hfill%
\panelvstack[0.18\textwidth][1mm]{C}%
  {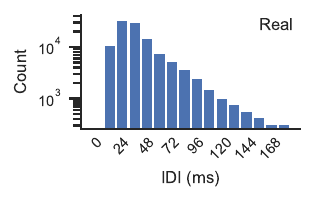}%
  {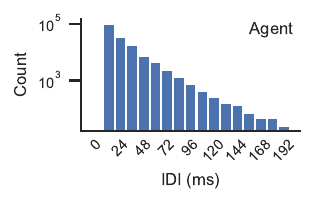}%
\panelvstack[0.18\textwidth][1mm]{D}%
  {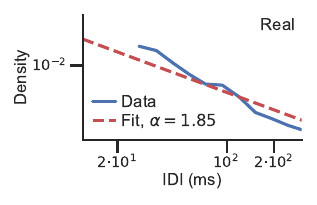}%
  {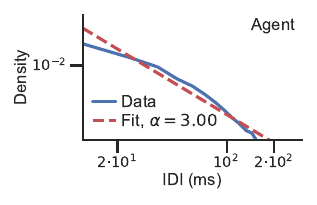}%

\vspace{1pt}

\panel{E}{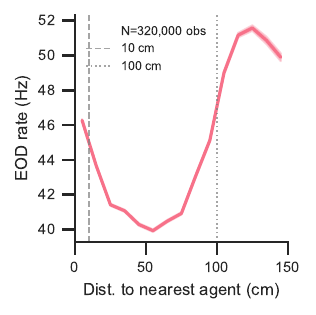}\hfill
\panel{F}{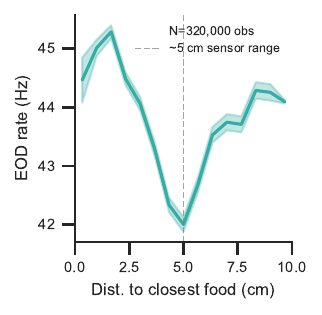}\hfill
\panel{G}{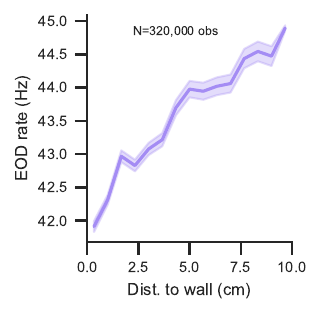}\hfill
\panel{H}{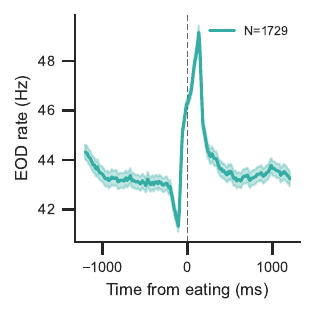}

\vspace{4pt}
\panel{I}{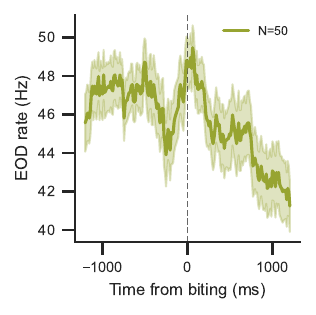}\hfill
\panel{J}{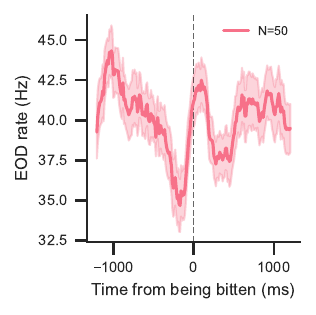}\hfill
\panel{K}{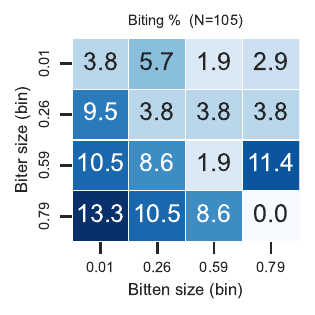}\hfill
\panel[\PanelW][3mm]{L}{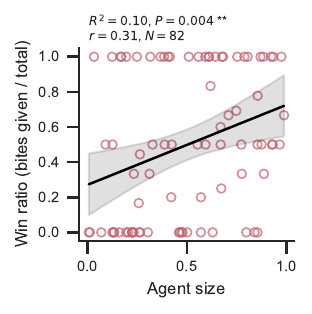}

\vspace{4pt}
\panel[\PanelW][3mm]{M}{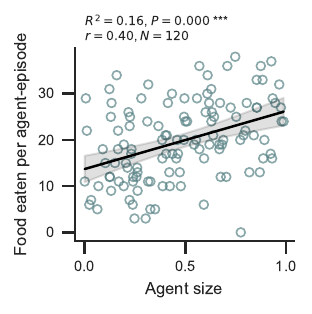}\hfill
\panel[\PanelW][3mm]{N}{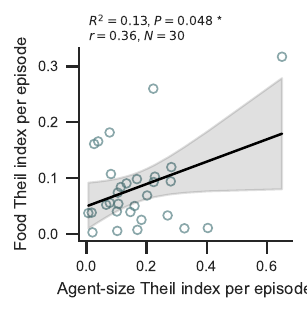}\hfill
\panel[\PanelW][3mm]{O}{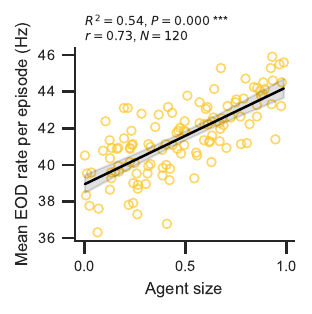}\hfill
\panel[\PanelW][3mm]{P}{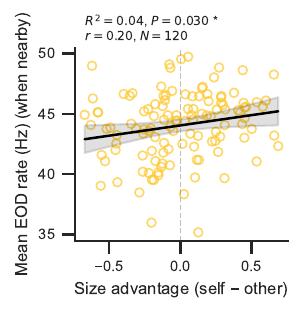}\hfill
\caption[]{
\textbf{Emergent social foraging in electrosensing agents.}
\textbf{(A)}
Ethogram from a representative multi-agent episode showing exploration, feeding, and social interactions;
thin blue bands indicate EOD rate for each agent.
\textbf{(B)}
Representative trajectories showing learned exploration and foraging in a shared arena with patchy food.
\textbf{(C)}
Inter-discharge interval (IDI) distributions for dyads of (top) real fish\citep{chrtkova2025unsupervised} and (bottom) trained agents, showing heavy-tailed pulse timing statistics in both.
\textbf{(D)}
Power-law fits to IDI distributions for real fish and trained agents.
\textbf{(E--G)}
EOD rate is modulated by spatial context, varying with distance to the nearest agent, food, and arena wall.
\textbf{(H--J)}
Peri-event EOD dynamics around eating, biting another agent, and being bitten, showing event-locked modulation of signaling; all curves show mean $\pm$ SEM.
\textbf{(K)}
Biting heatmap shows that larger agents tend to bite smaller agents.
\textbf{(L)}
Win ratio (fraction of bites given / total) increases with body size.
\textbf{(M)}
Larger agents consume more food per episode.
\textbf{(N)}
Food consumption inequality (Theil index) scales with body-size inequality across groups.
\textbf{(O)}
Body size correlates positively with EOD emission rate.
\textbf{(P)}
EOD rate advantage tracks size advantage between paired agents when in close range ($\leq 10$ cm).
}
\label{fig:behavior}
\end{figure*}

\subsection{\textbf{Trained agents develop flexible social foraging and context-dependent EOD modulation}}

Collective foraging in weakly electric fish involves both exploration and competition, with context-driven modulations of EOD signaling \citep{arnegardElectricOrganDischarge2005,von1992electrolocation,caputiElectricOrganDischarge1999}.
To test whether our agents reproduce these patterns, we evaluated them in arenas with either patchy or uniform food distributions (Fig.~\ref{fig:behavior}).
The multi-agent patchy panels used 70 cm $\times$ 70 cm arenas with four agents initialized with random positions, orientations, and size parameters, repeated for 30 episodes; these capture group-level foraging and dominance dynamics.
The inter-discharge interval (IDI), distance-modulation, and peri-event EOD panels used a two-agent wide assay in a 160 cm $\times$ 40 cm arena, repeated for 100 evaluation episodes; this narrower protocol isolates pairwise EOD dynamics with higher statistical power.

In the four-agent patchy assay, agents alternated between exploration, feeding, and social interactions, with context-dependent changes in EOD rates visible across agents (Fig.~\ref{fig:behavior}A,B).
In the two-agent wide assay, IDIs during foraging followed heavy-tailed distributions, as do those measured in real fish \citep{chrtkova2025unsupervised} (Fig.~\ref{fig:behavior}C); both admit power-law-like fits \citep{alstott2014powerlaw} over the same range (Fig.~\ref{fig:behavior}D), though the exponents do not quantitatively match.

Agents flexibly modulated EOD emission based on spatial context.
EOD rate varied systematically with distance to the nearest conspecific (Fig.~\ref{fig:behavior}E), distance to the closest food item (Fig.~\ref{fig:behavior}F), and distance to the nearest arena wall (Fig.~\ref{fig:behavior}G).
Peri-event analyses revealed event-locked modulation of signaling: EOD rate showed a sharp transient increase around eating events (Fig.~\ref{fig:behavior}H), a build-up preceding and decline following the agent biting another fish (Fig.~\ref{fig:behavior}I), and an elevation around times in which the agent was bitten (Fig.~\ref{fig:behavior}J).

Training with inter-agent heterogeneity led to emergent dominance-like asymmetries:
larger agents directed more biting toward smaller conspecifics, with win ratio increasing with body size (Fig.~\ref{fig:behavior}K,L).
Larger agents also consumed more food per episode (Fig.~\ref{fig:behavior}M),
and food-intake inequality scaled with body-size inequality across groups (Fig.~\ref{fig:behavior}N).
Body size tended to correlate positively with EOD emission rate (Fig.~\ref{fig:behavior}O),
and EOD rate advantage weakly tracked size advantage in close-range agent pairs (Fig.~\ref{fig:behavior}P).

Together, these results show that biophysically grounded sensory structure and simple individual incentives are sufficient for flexible, fish-like collective foraging across ecological conditions.

\subsection{\textbf{In silico perturbations reveal sensory and signaling contributions to collective foraging}}

\begin{figure*}[!tbp]
\centering

\panel[0.327\textwidth]{A}{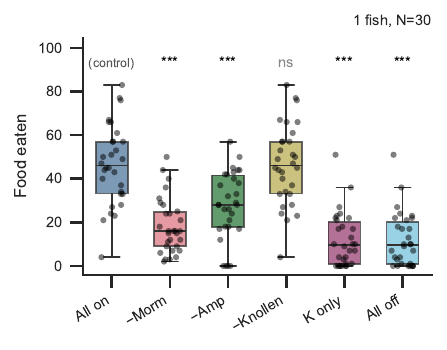}\hfill
\panel[0.327\textwidth]{B}{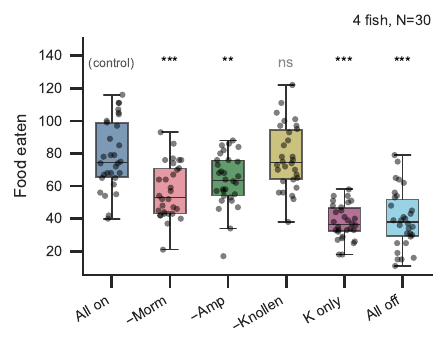}\hfill
\panel[0.327\textwidth]{C}{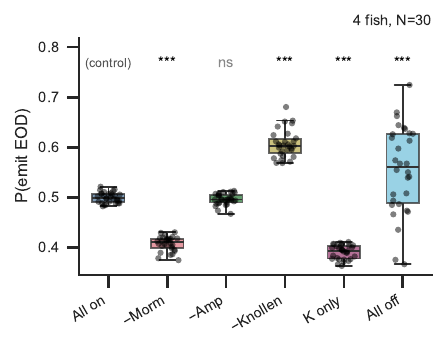}

\vspace{4pt}
\panel[0.327\textwidth]{D}{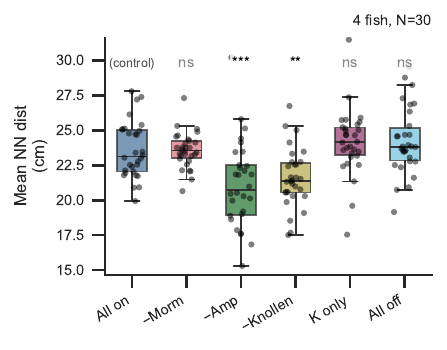}\hfill
\panel[0.327\textwidth]{E}{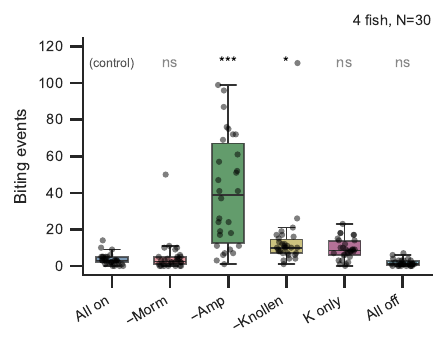}\hfill
\panel[0.327\textwidth]{F}{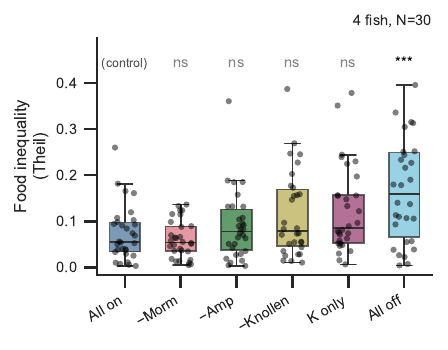}

\vspace{4pt}
\panel{G}{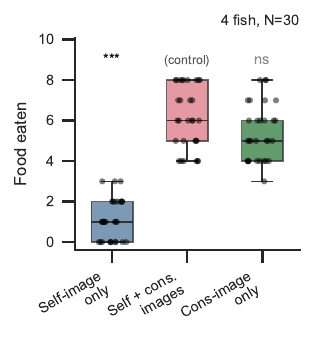}\hfill
\panel{H}{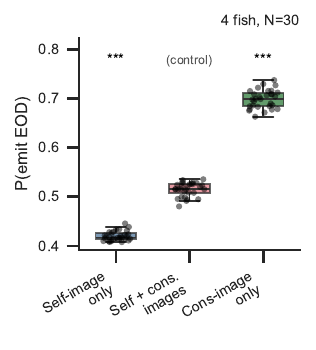}\hfill
\panel{I}{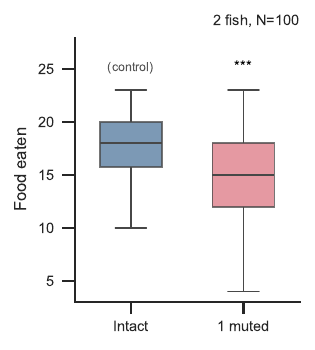}\hfill
\panel{J}{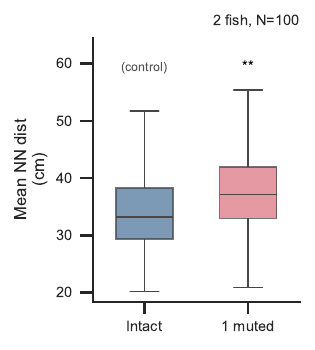}

\vspace{4pt}
\panel{K}{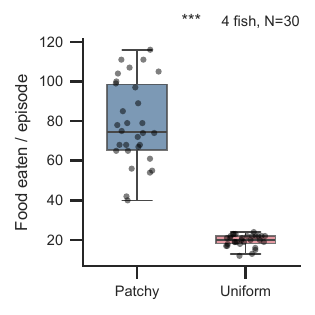}\hfill
\panel{L}{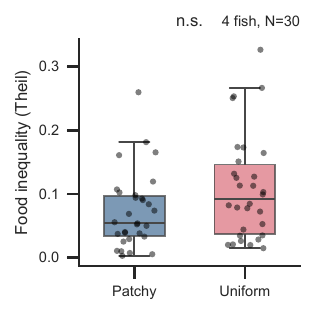}\hfill
\panel{M}{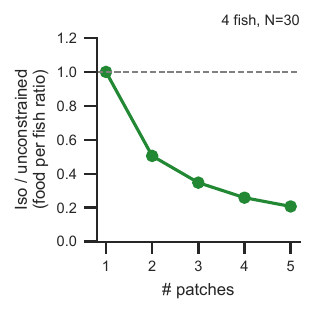}\hfill
\panel{N}{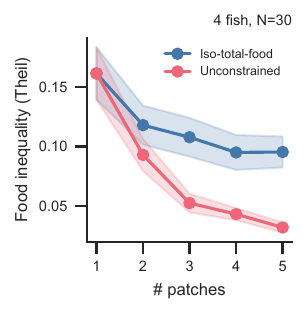}

\caption[]{
\textbf{In-silico interventions reveal potential causal factors underlying collective foraging.}
\textbf{(A--F)} Electroreceptor ablations compared against the full-sensor baseline using one-vs-control Dunnett tests.
\textbf{(A)} Food eaten by a single isolated agent under each sensor ablation; isolates foraging from social effects.
\textbf{(B)} Food eaten per agent in 4-fish groups under sensor ablation.
\textbf{(C)} EOD emission probability under sensor ablation.
\textbf{(D)} Mean nearest-neighbor spacing under sensor ablation.
\textbf{(E)} Number of biting events under sensor ablation.
\textbf{(F)} Food consumption inequality (Theil index) under sensor ablation.
\textbf{(G,H)} Collective sensing manipulations (self-EOD versus cons-EOD inputs to Mormyromast channel):
food eaten \textbf{(G)} and EOD emission probability \textbf{(H)}.
\textbf{(I,J)} EOD muting manipulations (intact group versus one-agent-muted group):
food eaten \textbf{(I)} and mean nearest-neighbor distance \textbf{(J)}.
\textbf{(K--N)} Environmental manipulations.
\textbf{(K,L)} Patchy versus uniform food distribution:
total food eaten \textbf{(K)} and food consumption inequality \textbf{(L)}.
\textbf{(M,N)} Iso-food versus unconstrained patch-number sweep (1--5 patches):
food-per-fish ratio of iso to unconstrained series \textbf{(M)} and food consumption (Theil) inequality \textbf{(N)} across patch counts.
Bar and point summaries show mean $\pm$ SEM where error bars are present; significance annotations are panel-level exploratory summaries.
}
\label{fig:ablations}
\end{figure*}

Virtual interventions make it possible to causally isolate mechanisms that are often difficult or impossible to manipulate experimentally. 
We therefore performed manipulations targeting electroreceptor channels, active signaling, and food availability, in four-fish groups (ablations and environmental), a two-agent assay (EOD muting), and a solo-fish control.

\textbf{Sensor ablations.}
We disabled each electroreceptor class in turn to identify causal contributions of short-range (Mormyromast, Ampullary) and long-range (Knollenorgan) channels (Fig.~\ref{fig:ablations}A--F).
Short-range sensor ablations reduced food consumption in both the solo assay (Fig.~\ref{fig:ablations}A, single agent without conspecifics) and four-agent groups (Fig.~\ref{fig:ablations}B), indicating effects on individual foraging that persist independently of social interaction.
Knollenorgan ablation left total food consumption intact but reshaped social organization: EOD emission probability and biting rose while inter-agent spacing fell (Fig.~\ref{fig:ablations}C--F).

\textbf{Collective sensing.}
Within the Mormyromast channel, we independently gated self- and conspecific-EOD inputs (Fig.~\ref{fig:ablations}G,H).
Removing the conspecific-EOD image (self-image-only condition) drastically reduced food consumed (Fig.~\ref{fig:ablations}G), while removing the self-EOD image (cons-image-only condition) did not show a significant effect on foraging.
EOD emission probability was lower in self-image-only agents and higher in cons-image-only agents (Fig.~\ref{fig:ablations}H).

\textbf{EOD muting.}
Silencing EOD emission in one of two agents reduced total food consumption and increased mean nearest-neighbor distance (Fig.~\ref{fig:ablations}I,J), indicating that active electrocommunication supports both effective foraging and social spacing.

\textbf{Environmental manipulations.}
Patchy food distributions yielded substantially more food per episode than uniform layouts (Fig.~\ref{fig:ablations}K), with no significant difference in consumption inequality between the two conditions (Fig.~\ref{fig:ablations}L).
We swept patch count from 1 to 5 under two conditions: an iso-food series (total food held constant, food-per-patch decreasing) and an unconstrained series (food-per-patch held constant, total food increasing).
Per-fish intake fell progressively in the iso-food series relative to the unconstrained series (Fig.~\ref{fig:ablations}M); inequality fell with increasing patch count in both series, somewhat more steeply in the unconstrained condition (Fig.~\ref{fig:ablations}N).

Together, these manipulations identify distinct causal roles for short-range channels (foraging), Knollenorgan sensing (social spacing), collective sensing, and resource structure (inequality), yielding falsifiable predictions for ablation and competition experiments in real fish.

\begin{figure*}[!tbp]
\centering

\panel{A}{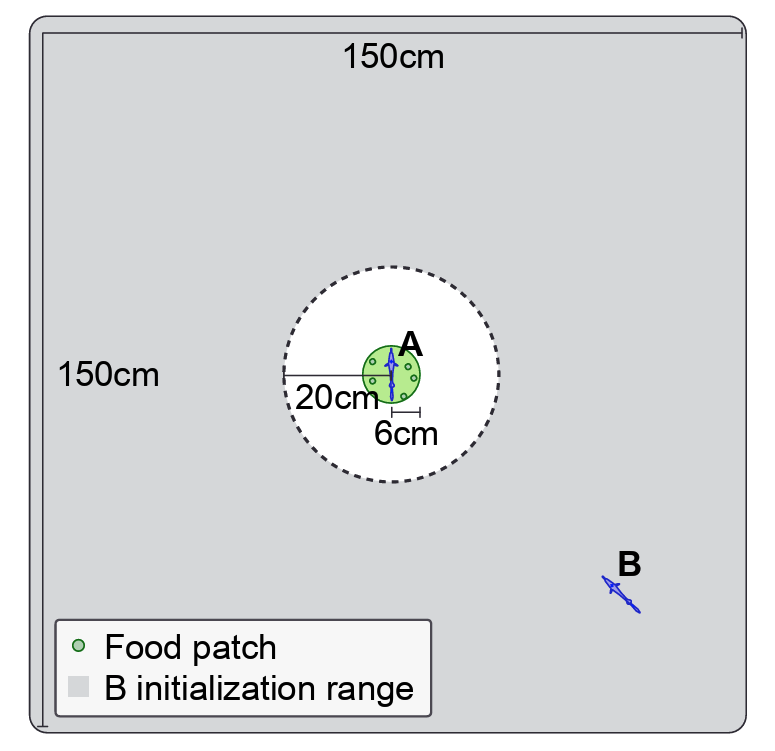}\hfill
\panel{B}{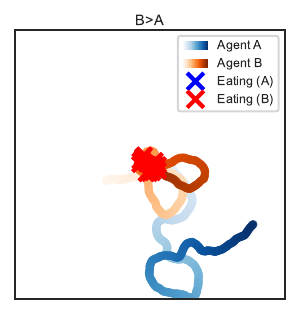}\hfill
\panel{C}{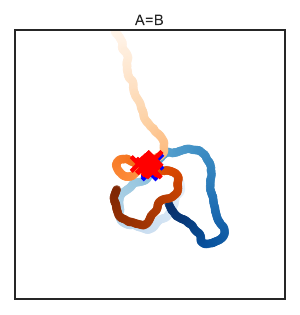}\hfill
\panel{D}{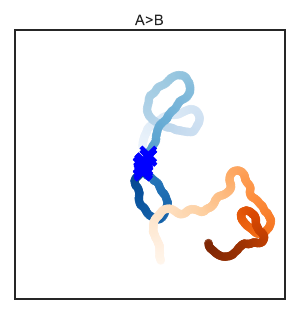}

\vspace{4pt}
\panel{E}{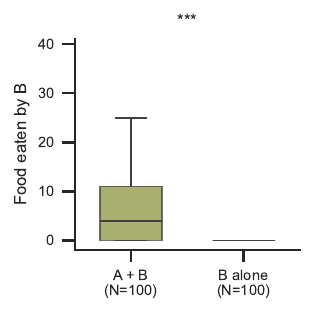}\hfill
\panel{F}{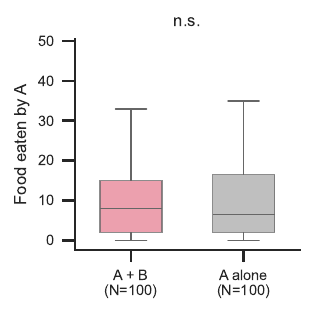}\hfill
\panel{G}{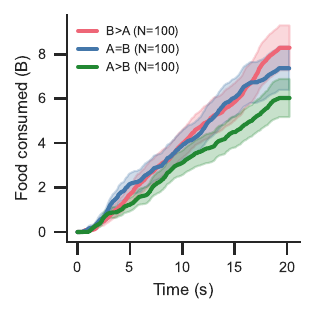}\hfill
\panel{H}{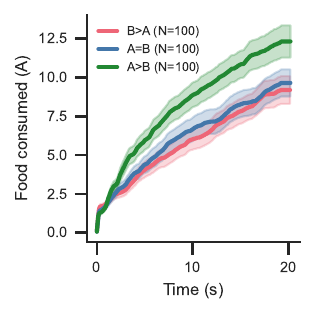}\hfill

\vspace{4pt}
\panel{I}{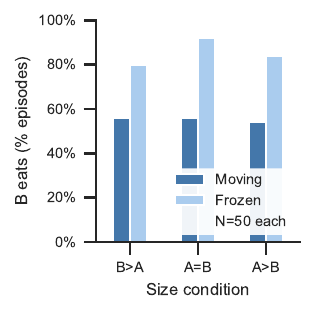}\hfill
\panel{J}{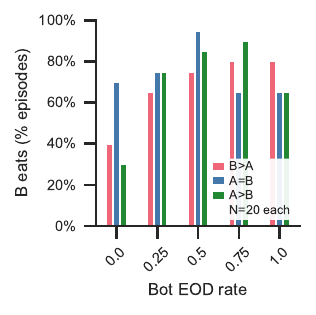}\hfill
\panel{K}{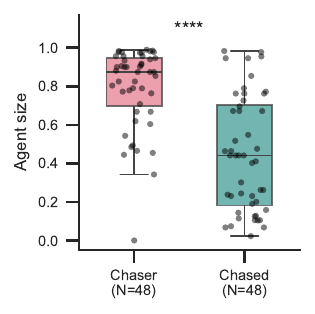}\hfill
\panel{L}{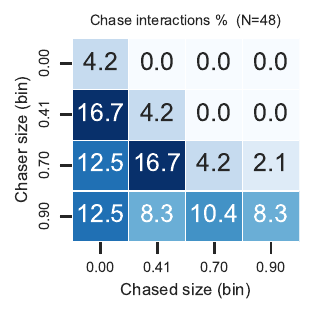}\hfill

\vspace{4pt}
\panel{M}{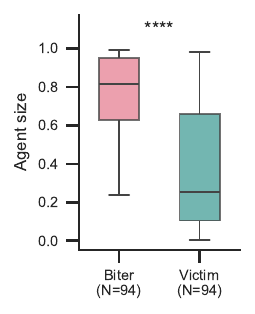}\hfill
\panel{N}{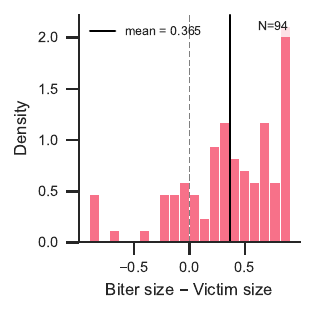}\hfill
\panel{O}{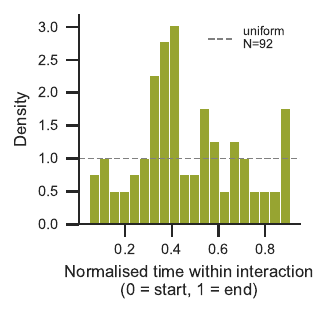}\hfill 
\panel{P}{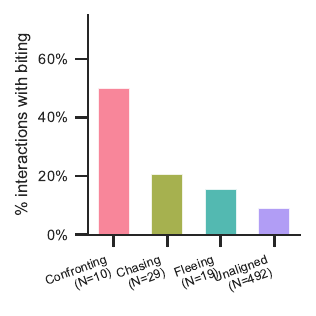}\hfill


\caption[]{
\textbf{Two-fish foraging assays reveal the role of signaling and dominance in social interactions.}
\textbf{(A)}
Minimal two-agent single-patch foraging assay.
Agent A is initialized within a replenishing food patch, while agent B is initialized at a random location within communication range.
\textbf{(B--D)}
Example trajectories showing B approaching a patch occupied by A;
trajectories are colored from light (start) to dark (end).
\textbf{(B)}
When B is larger than A, B approaches the patch directly and displaces A.
\textbf{(C)}
When A and B are the same size, both agents access the patch.
\textbf{(D)}
When B is smaller than A, B's approach is more indirect and patch access is reduced.
\textbf{(E,F)}
Food consumed by B \textbf{(E)} and A \textbf{(F)} in social versus non-social (alone) conditions across size asymmetry classes.
\textbf{(G)}
Timecourse of food consumption by agent B across size asymmetry classes; curves show mean $\pm$ SEM.
\textbf{(H)}
Timecourse of food consumption by agent A across size asymmetry classes; curves show mean $\pm$ SEM.
\textbf{(I,J)}
Agent A is replaced by a rule-based random walker "bot" confined to the central patch.
\textbf{(I)}
B's patch-reach rate grouped by relative size, across all bot EOD rates, stratified by whether bot is moving or frozen in place.
\textbf{(J)}
B's patch-reach rate grouped by walker EOD emission rate, stratified by size condition and aggregated across bot-movement conditions.
\textbf{(K,L)}
In arenas with two agents and uniformly distributed food, larger agents are more likely to act as chasers:
body size by chaser/chased role as a boxplot \textbf{(K)} and as a pairwise size heatmap \textbf{(L)}.
\textbf{(M)}
Body size by biter/bitten role (boxplot); larger agents bite more frequently.
\textbf{(N)}
Size advantage of the biter over the bitten agent.
\textbf{(O)}
Timing of biting events within interaction windows.
\textbf{(P)}
Biting rate broken down by interaction class.
Boxplot significance annotations use panel-level exploratory Mann-Whitney or Wilcoxon tests as appropriate.
}
\label{fig:twofish}
\end{figure*}

\subsection{\textbf{Dyadic assays reveal size and signal-dependent effects on foraging and interactions}}

Two-fish assays ($150$ cm $\times 150$ cm) allow closer inspection of how EOD modulation and relative-size asymmetries shape competition at the smallest social scale.
We evaluated pairs of agents in a one-patch dyadic task, with resident A initialized on a replenishing patch and intruder B initialized away from it, repeating 100 trials for each size condition (Fig.~\ref{fig:twofish}A).

Example trajectories illustrate how relative size shaped approach behavior: dominant intruders approached the patch directly and displaced the resident, same-size pairs often shared access, and subordinate intruders approached more indirectly with reduced patch access (Fig.~\ref{fig:twofish}B--D).
Intruder B consumed significantly more food when A was present than when foraging alone (Fig.~\ref{fig:twofish}E), while resident A's total food intake was not significantly affected by B's presence (Fig.~\ref{fig:twofish}F).
Consumption timecourses revealed a graded size-asymmetry effect: 
B accumulated food faster when dominant (B$>$A) and slower when subordinate (A$>$B), with A showing the complementary pattern (Fig.~\ref{fig:twofish}G,H).

Replacing the resident with a rule-based patch-confined random walker ("bot") isolated the contribution of EOD rate and movement state.
B reached the patch more often when the bot was frozen than when moving (Fig.~\ref{fig:twofish}I).
B's patch-reach rate was generally higher when the bot emitted EODs at any nonzero rate than when silent, with success peaking at intermediate emission rates (Fig.~\ref{fig:twofish}J).

In separate two-agent square arenas (70 cm $\times$ 70 cm) with uniform food distributions, larger agents were more likely to act as chasers and smaller ones as chased (Fig.~\ref{fig:twofish}K,L).
Similarly, biters tended to be larger than their victims (Fig.~\ref{fig:twofish}M,N).
Biting was concentrated in the first half of interaction windows, peaking at approximately 40\% of the normalized interaction duration (Fig.~\ref{fig:twofish}O).
Confronting interactions had the highest biting rate (${\approx}50\%$), followed by chasing, fleeing, and unaligned interactions (Fig.~\ref{fig:twofish}P) (See Methods for interaction definitions).

These assays identify relative body size and active EOD signaling as key determinants of competitive outcomes in dyadic interaction.

\begin{figure*}[!tbp]
\centering
\panel{A}{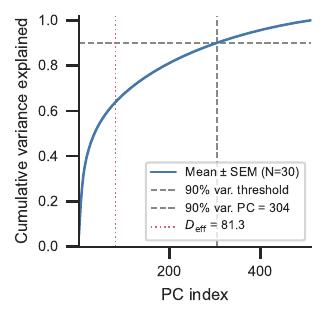}\hfill
\panel{B}{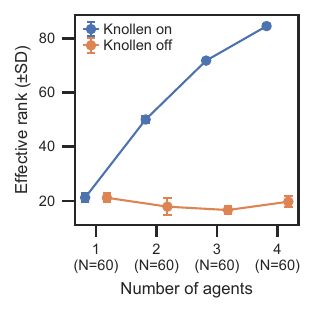}\hfill
\panel{C}{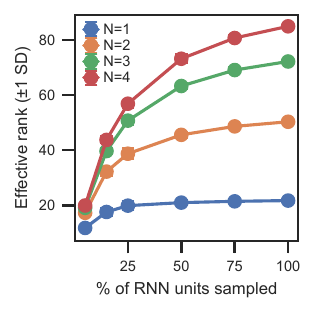}\hfill
\panel{D}{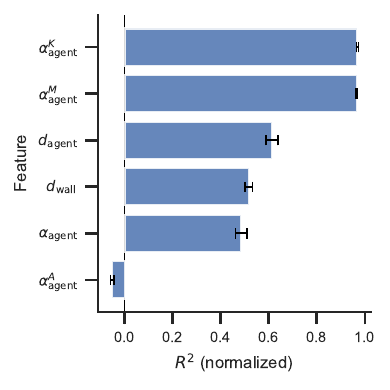}

\vspace{4pt}
\panel[0.48\textwidth]{E}{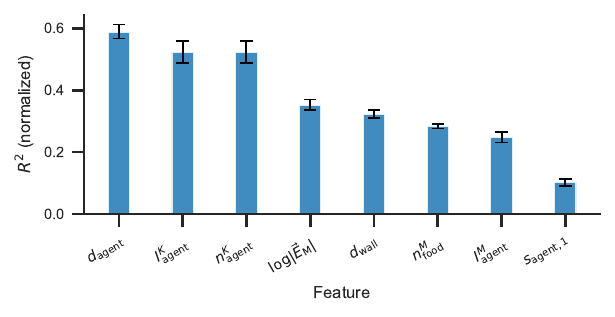}
\panel[0.48\textwidth]{F}{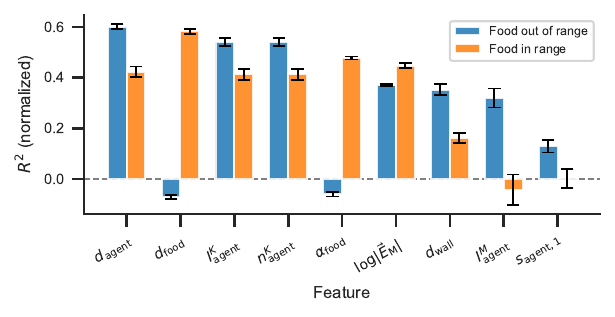}
\vspace{4pt}
\panel{G}{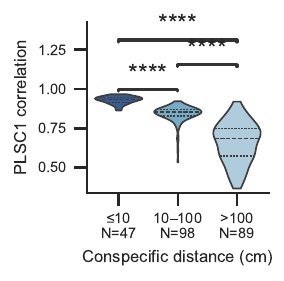}\hfill
\panel{H}{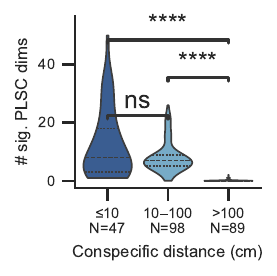}\hfill
\panel{I}{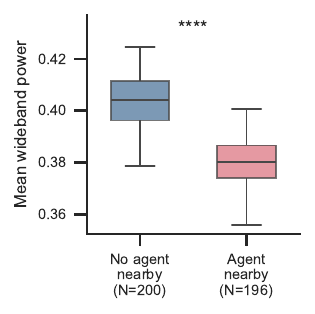}\hfill
\panel{J}{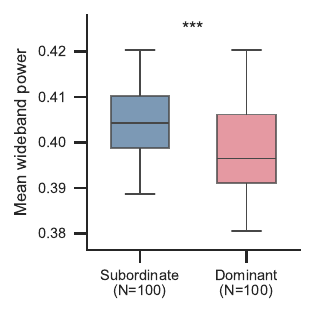}\hfill

\vspace{4pt}
\caption[]{
\textbf{Recurrent network activity encodes task, social, and behavioral context.}
\textbf{(A)}
PCA of recurrent hidden-state activity shows that most variance is captured by a relatively low-dimensional subspace, with effective rank well below the full hidden-state size ($D_H=512$).
\textbf{(B)}
Effective dimensionality (mean $\pm$ SD) increases with the number of interacting agents. 
When Knollenorgan is ablated, it stays steady around the effective dimensionality of the solo foraging setting.
\textbf{(C)}
Subsampling analysis shows that estimated effective rank increases with the fraction of recurrent units sampled, across group sizes, but even a small subsampling fraction can reveal differences across group sizes.
\textbf{(D)}
Homing-task linear decoding shows that the Knollenorgan-defined error angle ($\Theta^K_{\mathrm{error}}$) is decoded with higher test performance than the straight-line agent angle ($\Theta_{\mathrm{agent}}$), indicating stronger representation of electrosensory task geometry in recurrent activity.
\textbf{(E)}
Linear decoding performance for task-relevant variables across all timesteps (combined condition): nearest-agent distance ($d_\mathrm{agent}$), Knollenorgan presence and count ($I^K_\mathrm{agent}$, $n^K_\mathrm{agent}$), log Mormyromast field magnitude ($\log|\vec{E}_\mathrm{M}|$), and wall distance ($d_\mathrm{wall}$).
\textbf{(F)}
Linear decoding performance split by food-proximity condition (food near vs.\ far): distance to conspecific ($d_\mathrm{agent}$), food distance ($d_\mathrm{food}$), food bearing ($\alpha_\mathrm{food}$), and Knollenorgan features ($I^K_\mathrm{agent}$, $n^K_\mathrm{agent}$); error bars show SEM across grouped cross-validation folds for both \textbf{(E--F)}.
\textbf{(G,H)}
Partial least-squares correlation (PLSC) analysis: 
first-component correlation \textbf{(G)} and number of statistically significant shared dimensions \textbf{(H)} are both higher for agent pairs within interaction range than for out-of-range pairs; PLSC significance uses the circular-shift null described in Methods.
\textbf{(I)}
Episode-level mean wideband RNN power is lower for timesteps when a conspecific is nearby.
\textbf{(J)}
Episode-level mean wideband RNN power differs by relative size, with subordinate (smaller) agents showing higher wideband power than dominant (larger) agents.
}
\label{fig:rnn}
\end{figure*}

\subsection{\textbf{Recurrent dynamics encode task variables and social interaction state}}

Principal component analysis (PCA) of four-agent patchy foraging episodes yielded an effective rank of $D_\mathrm{eff} = 81.3$ (304 PCs for 90\% variance; Fig.~\ref{fig:rnn}A), well below the $D_H = 512$ nominal capacity.

Effective dimensionality scaled with group size when the Knollenorgan was intact, but remained flat near the solo baseline when the Knollenorgan was ablated (Fig.~\ref{fig:rnn}B), implicating long-range signaling as the primary driver of social-context-dependent dimensionality expansion.
A subsampling analysis confirmed that the group-size scaling is visible even with partial unit recordings (Fig.~\ref{fig:rnn}C).

In the homing task, a linear decoder trained on RNN hidden states recovered the Knollenorgan-defined error angle ($\theta^K_\mathrm{error}$) with substantially higher performance than the straight-line angle to the target ($\theta_\mathrm{agent}$; Fig.~\ref{fig:rnn}D), indicating that the network represents field-line geometry rather than Euclidean geometry.

In two-agent wide-assay episodes, collapsed across all timesteps, nearest-agent distance ($d_\mathrm{agent}$) and Knollenorgan features (presence $I^K_\mathrm{agent}$; count $n^K_\mathrm{agent}$) were the top-decoded variables, followed by log Mormyromast field magnitude ($\log|\vec{E}_\mathrm{M}|$) and wall distance ($d_\mathrm{wall}$) (Fig.~\ref{fig:rnn}E).
Furthermore, decoding revealed strong context-dependence (Fig.~\ref{fig:rnn}F):
distance to the conspecific ($d_\mathrm{agent}$) was reliably decoded across both food-range conditions;
food distance ($d_\mathrm{food}$) and food bearing ($\alpha_\mathrm{food}$) tended to improve in decoding performance when food was within sensing range.
Knollenorgan features ($I^K_\mathrm{agent}$, $n^K_\mathrm{agent}$) decoded at moderate levels in both conditions.

Partial least squares correlation (PLSC) revealed proximity-dependent correlated latent structure between paired agents (Fig.~\ref{fig:rnn}G,H).
First-component correlation decreased across conspecific-distance bins (${\leq}10$, $10$--$100$, and ${>}100$~cm; Fig.~\ref{fig:rnn}G).
The number of statistically significant shared dimensions was similar for pairs within 100~cm (n.s.\ between the ${\leq}10$ and $10$--$100$~cm bins) but collapsed to near zero beyond 100~cm (Fig.~\ref{fig:rnn}H).

RNN power-spectrum analyses showed that the presence of a nearby conspecific significantly reduced mean wideband RNN power (Fig.~\ref{fig:rnn}I),
while subordinate (smaller) agents carried significantly higher mean wideband power than dominant (larger) agents (Fig.~\ref{fig:rnn}J).

Together, these analyses show that trained agents encode electrosensory task geometry and social context in structured, low-dimensional population dynamics, and that interacting agents exhibit proximity-dependent correlated latent dynamics.


\section{Discussion}
Our results demonstrate that biophysically grounded agents trained with multi-agent reinforcement learning can qualitatively reproduce a broad range of known weakly electric fish behaviors and yield causal insights that would be difficult to obtain experimentally.
By validating against established experiments, extending to novel conditions, and conducting interventions that are otherwise impractical, the framework provides a useful complement to empirical research.
It shows that biophysically grounded sensory structure, coupled with simple individual reward structures, is sufficient for rich, flexible group dynamics that mirror natural behavior.
Our goal is not to provide a complete biophysical replica of weakly electric fish, nor to claim that the mechanisms discovered here are uniquely required in biological animals;
rather, we use a deliberately simplified but biologically grounded closed-loop model to test whether realistic electrosensory structure, individual incentives, and multi-agent interaction are \textit{sufficient} to generate observed collective phenomena like electrosensing, foraging, and social dynamics.

\subsection{Connections with known biology}

Our agents reproduced a range of well-characterized behaviors of weakly electric fish, providing qualitative validation of the framework and suggesting mechanistic hypotheses for phenomena that remain difficult to study experimentally.
Furthermore, virtual experiments in Figs. \ref{fig:ablations}, ~\ref{fig:twofish} and \ref{fig:rnn} provide testable hypotheses that motivate future experimental follow-up.

\textit{Curvilinear homing trajectories.}
Without explicitly encoding homing geometry, agents followed smooth curved paths to targets, replicating the hallmark behavior reported in classic experiments on real fish \citep{hopkins1997quantitative,von1999active} (Fig.~\ref{fig:homing}).
Agents using a non-electric ray-casting baseline, which provided direct distance and object-type cues, followed straight-line paths instead (Fig.~\ref{fig:homing}B), confirming that curved trajectories arise from the electrosensory modality rather than from goal-directed navigation in general.
RNN decoding further showed that the network represents field-line geometry rather than Euclidean geometry (Fig.~\ref{fig:rnn}D), consistent with the electrosensory basis of the curvilinear trajectories.

\textit{Heavy-tailed and context-dependent EOD statistics.}
IDI histograms during foraging were heavy-tailed and context-dependent, qualitatively matching distributions measured in Mormyrid fish in laboratory and field settings \citep{mollerElectricOrganDischarge1989,carlsonStereotypedTemporalPatterns2004,schusterCountSparkEcho2001,henninger2018statistics,gebhardtElectricDischargePatterns2012}.
Our results suggest that this statistical structure need not be hard-wired; it can emerge from a reward-driven policy that flexibly modulates EOD emission based on foraging state, social context, and prior history.
Event-locked analyses further revealed EOD modulation around eating, biting, and being-bitten events (Fig.~\ref{fig:behavior}H--J), rather than emitting at a fixed baseline rate.
While we do not see a full cessation of EOD emission by subordinate individuals as observed in real fish \citep{mollerElectricFishes1995,carlsonElectricSignalingBehavior2002}, we do see a size-dependent reduction in EOD emission rate during interactions (Fig.~\ref{fig:behavior}P).

\textit{Emergent foraging dynamics.}
Agents trained under simple individual-fitness incentives spontaneously developed collective foraging strategies that varied flexibly with food distribution, group composition, and social context (Fig.~\ref{fig:behavior}). 
Agents demonstrated behavioral flexibility by alternating between exploration, targeted approach, and social interaction in ways consistent with observations of Mormyrid foraging \citep{friedman_tracking_1996,biswas2023mode}, and with experimental evidence that social interactions improve foraging efficiency and compress income inequality in fish groups \citep{harpazSocialInteractionsDrive2020}. 
This suggests that flexible collective foraging can emerge from the alignment of individual incentives with shared sensory ecology, without explicit group-level coordination signals.

\textit{Sensory contributions.}
Mirroring prior experimental work \citep{emde1998finding}, ablating either short-range sensor (Mormyromasts or Ampullary) reduces foraging rate but does not stop foraging altogether (Fig.~\ref{fig:ablations}A--B).

\textit{Collective sensing.}
Within the Mormyromast channel, removing access to the conspecific-EOD image (self-image-only condition) drastically reduced food consumption, while removing the self-EOD image (cons-image-only condition) had little effect (Fig.~\ref{fig:ablations}G) \citep{pedrajaCollectiveSensingElectric2024,wallachInternalModelCanceling2023,enikolopovInternallyGeneratedPredictions2018}.
EOD emission probability was correspondingly lower in agents restricted to the self-generated image and elevated in those restricted to the conspecific image (Fig.~\ref{fig:ablations}H), implicating the cons-EOD image in both foraging efficiency and signaling regulation.

\textit{Long-range signaling.}
Classic \textit{producer}-\textit{scrounger} theory in behavioral ecology \citep{giraldeau2018social} tells us that individuals can exploit conspecific signals as public information at low cost, providing a functional basis for long-range sensory tuning.
In our agents, Knollenorgan ablation left total food consumption largely intact while increasing EOD emission probability and close-range aggression, and reducing inter-agent spacing (Fig.~\ref{fig:ablations}C--F) \citep{pedrajaCollectiveSensingElectric2024}, suggesting this channel contributes primarily to social coordination rather than individual foraging efficiency.
This generates a testable prediction:
selectively blocking Knollenorgan-mediated signaling in real fish groups should increase agonistic interactions and reduce social spacing without substantially impairing individual foraging efficiency.

\textit{Active signaling.}
Silencing a single agent's EOD emission reduced group-level food consumption and increased inter-agent spacing (Fig.~\ref{fig:ablations}I,J), indicating that active electrocommunication supports both foraging coordination and competitive spacing, distinct from the passive-reception role of the Knollenorgans \citep{arnegardElectricOrganDischarge2005}.

\textit{Environmental structure.}
Patchy food distributions supported higher per-episode consumption than uniform distributions (Fig.~\ref{fig:ablations}K) without significantly affecting inequality (Fig.~\ref{fig:ablations}L), suggesting that spatial structure boosts collective foraging efficiency without systematically concentrating resources.
Increasing patch count with fixed total food progressively reduced per-fish intake while decreasing inequality (Fig.~\ref{fig:ablations}M,N), consistent with the idea that resource subdivision reduces competitive skew by creating multiple simultaneous access points.

\textit{Aggression and size-dependent asymmetries.}
Training agents with size and speed heterogeneity produced emergent dominance-like asymmetries: 
larger agents consumed disproportionately more food and directed more aggression toward smaller conspecifics, consistent with ethological observations of size- and EOD-related dominance in weakly electric fish \citep{arnegardElectricOrganDischarge2005,terleph2003effects}.
The two-fish assays further showed that larger intruders were more likely to displace residents and monopolize food patches, mirroring behavioral observations of dyadic agonistic encounters in Mormyrids and size-dependent dominance dynamics in other weakly electric species \citep{terleph2003effects,worm2018evidence,o2024dynamics}.
Notably, dominance-like behavior emerged from size-dependent movement constraints and asymmetric penalties for being bitten, without any explicit reward encouraging biting or chasing.
In real weakly electric fish, subordinate individuals sometimes temporarily silence their EODs during agonistic encounters as a submissive signal \citep{mollerElectricFishes1995}; whether a similar learned suppression emerges in agents trained under stronger social pressure or with richer communication structure is a direct test for future work.
These assays generate testable predictions: larger intruders should systematically displace smaller residents, with displacement probability scaling with size difference; and intruders should reach occupied patches more readily when the resident is stationary or actively emitting, consistent with conspecific EODs serving as informative cues for patch discovery \citep{chrtkova2025unsupervised} (Fig.~\ref{fig:twofish}A--J).

\textit{Neural dynamics and correlated recurrent-state structure.}
Effective dimensionality of population activity increased with the number of interacting agents, and this scaling was abolished when the Knollenorgan was ablated (Fig.~\ref{fig:rnn}B,C), implicating long-range electrocommunication as a driver of social-context-dependent state-space expansion.
PLSC analysis showed that the first shared axis between two agents' RNN states was significantly more correlated at close range than when agents were out of range, with shared dimensionality collapsing to near zero beyond 100~cm, showing a proximity-dependent pattern that qualitatively parallels the shared neural subspaces recently reported in interacting mice \citep{zhang2025inter,redman2026predictive}.
Wideband RNN power was suppressed when a conspecific was nearby (Fig.~\ref{fig:rnn}I) and was elevated in subordinate relative to dominant agents (Fig.~\ref{fig:rnn}J), indicating proximity- and status-dependent modulation of recurrent dynamics.

\subsection{Limitations and future work}
We do not collect new experimental behavioral or neural data here.
The model is instead intended to precede and guide future data collection by identifying regimes, perturbations, and readouts most informative to test experimentally.
This is especially important in weakly electric fish collectives, where simultaneous behavioral tracking, EOD assignment, and multi-animal neural recordings remain technically challenging \citep{zheng2025keypoint}.
Our framework provides a controlled setting in which signal sources, sensory access, neural states, and perturbations are fully known, allowing us to generate experimentally actionable predictions before committing to costly animal experiments.

Future work will use this framework in closer dialogue with experimental data.
Beyond foraging, we will train agents on richer social tasks such as territorial defense, competition, and cooperation, and test whether the resulting EOD sequences and interaction motifs match newly collected behavioral and neural data \citep{huang2025inputdsa}.
A further direction is to probe context-modulated electrocommunication through unsupervised machine-translation approaches \citep{singh2025proposal}, where virtual agents provide a fully observable ground truth for signal-state-action relationships while biological experiments test whether similar structure exists in real fish.

A broader limitation of RL/MARL agents is that out-of-distribution generalization can be brittle.
Our training regime partially mitigates this by exposing agents to substantial variability across initial positions, orientations, food configurations, social contexts, and stochastic rollouts, encouraging policies that generalize across a broad family of parameter regimes rather than memorizing a single arena configuration.

The model makes several simplifying assumptions.
Agents are modeled as circular particles rather than elongated, three-dimensional fish bodies.
The simulation does not model fluid dynamics or fluid-body interactions, and the electric field and sensory models assume electrostatic conditions in a time-discretized simulation with a 12 ms timestep, ignoring within-step electrodynamics.
Since real weakly electric fish continuously integrate information across electrosensory, visual, and mechanosensory modalities, behaviors observed in our agents may differ quantitatively from those in real collectives, particularly at short range where hydrodynamic cues dominate.
We do not model sex differences, which in real weakly electric fish drive rich EOD waveform variation and mate-preference dynamics \citep{carlsonElectricSignalingBehavior2002,zhou2006structure}.
Future work could address these limitations through embodied 3D simulation with fluid dynamics, richer contact mechanics for aggression, sexual dimorphism, longer-timescale training, and larger arenas supporting more naturalistic exploration.
Regardless, the framework demonstrates that virtual experiments can narrow the hypothesis space, accelerate experimental design, and uncover principles of electrosensory collective behavior that would be difficult or impossible to access through conventional experiments alone.

\section{Methods}

\subsection{Simulation details}
Simulations consisted of $N$ artificial weakly electric fish (WEF)-like agents, typically $N=4$, interacting with each other and with simulated prey in a bounded 2D arena. 
At each simulation step, neural-network policies generated actions for each agent, and the simulator updated agent kinematics, collisions, prey interactions, EOD generation, electric-field interactions, and sensor observations.

The simulator used first-order command kinematics: policy outputs were mapped directly to bounded translational and rotational velocity commands. The simulator enforced non-overlap constraints and wall collisions, and ran at 83 frames per second to match the observed latency in the WEF's social echo response (12 ms) \citep{heiligenberg2012principles}, close to the minimum observed inter-pulse interval in this WEF species.

\subsection{Agent design}
Each agent was represented as a circular particle with diameter $d$, allocentric (ground-frame) orientation $\theta$, and allocentric position $\mathbf{r} = (x, y)$.

Agents chose actions independently at every timestep from a multivariate action space comprising:
(1) forward translation,
(2) angular rotation,
(3) electric organ discharge (EOD) emission, and
(4) bite.
These collectively controlled locomotion, electrolocation, communication, and close-range aggression.
The policy output was continuous; the environment interpreted movement outputs as velocity commands and thresholded the EOD and bite channels into binary events.

Observations included input from three types of biomimetic electrosensors, last action taken, and egocentric displacement vectors. 
We introduced inter-agent heterogeneity by assigning each agent a scalar ``size'' parameter, which scaled its maximum speed.  
An agent's own size was included as a fixed value context input in the observation vector to allow size-dependent behavioral variation.

\subsection{Electric field model}
We followed the electrosensing model introduced by Chen et al.\ \citep{chenModelingSignalBackground2005} and built on a broader literature on electrosensory processing \citep{enikolopovInternallyGeneratedPredictions2018,mullerContinualLearningMultiLayer2019,wallachInternalModelCanceling2023}.

\subsubsection{Field generation}
Electric fields entered the simulation through two primary source classes. 
First, agents generated actively controlled monopole pairs that approximated pulsatile electric discharges from fish electric organs. 
Second, agents and prey carried always-on dipoles representing intrinsic electric sources.

In our agents, the electric organ was modeled as two internal monopoles with equal and opposite source strengths $\pm q$ (in coulombs) positioned at $\pm d/4$ along the body axis, avoiding singularities at sensor locations.
The total electric field $\mathbf{E}_{generated}$ at position $\mathbf{r}$ combines monopole and dipole contributions:
\begin{equation*}
    \mathbf{E}_{generated} = \mathbf{E}_{active} + \mathbf{E}_{intrinsic}
\end{equation*}

\begin{equation}
    \mathbf{E}_{active}(\mathbf{r}) = \frac{1}{4\pi\epsilon_0}\sum_i \frac{q_i(\mathbf{r}-\mathbf{r}_i)}{|\mathbf{r}-\mathbf{r}_i|^3}
\end{equation}

\begin{equation}
    \mathbf{E}_{intrinsic}(\mathbf{r}) = \frac{1}{4\pi\epsilon_0}\sum_i \frac{3(\mathbf{p}_i\cdot\hat{\mathbf{r}}_i)\hat{\mathbf{r}}_i - \mathbf{p}_i}{|\mathbf{r}-\mathbf{r}_i|^3}
\end{equation}

where $\epsilon_0 = 8.854\times10^{-12}$ F/m and $\hat{\mathbf{r}}_i$ is the unit vector from source $i$ to the measurement point.

\subsubsection{Field interactions with prey and agents}

Prey and agent particles were modeled as conducting spheres on which dipole moments were induced by the external electric field.
The induced dipole moment is given by:
\begin{equation}
    \mathbf{p}_{induced} = 4\pi\epsilon_0 R^3 \chi \mathbf{E}_{external}
\end{equation}
where 
$\chi = \left(\frac{\epsilon / \epsilon_0 - 1}{\epsilon / \epsilon_0 + 2}\right)$ ($\chi=-0.5$ per Chen et al 2005 \citep{chenModelingSignalBackground2005}),
$\epsilon_0$ is the permittivity of free space,
$\epsilon$ is the permittivity of the dielectric material,
$R$ is the radius of the sphere,
$\mathbf{E}_{\text{external}}$ is the total external electric field intersecting the object at its location.
The negative sign signifies that the induced dipole is oriented opposite to the external field.

\subsubsection{Field interactions with walls}
The arena walls were modeled as non-conducting boundaries, and we used the method of images \citep{griffiths2023introduction} to account for their effect on the electric field. 
This method introduced virtual image sources outside the domain to enforce the field boundary conditions.
We reflected all active and induced sources and dipoles off each of the arena walls one time, resulting in images at the same distance from the boundary on the other side and with the same magnitude and sign (no sign-flip for non-conducting boundaries \citep{griffiths2023introduction}); this is a first-order image approximation, with higher-order reflections neglected.

\subsection{Electroreception}
The combined field available to sensors was:
\begin{equation*}
    \mathbf{E}_{total} = \mathbf{E}_{active} + \mathbf{E}_{intrinsic} + \mathbf{E}_{induced} + \mathbf{E}_{reflected} 
\end{equation*}

Agents sensed this field through three biomimetic receptor classes: Mormyromasts, Knollenorgans, and Ampullary sensors.
Sensor types were chosen to approximate the functional specialization, spatial arrangement, and relative receptor counts of real mormyrid electroreceptors.

\subsubsection{Mormyromast sensors}
These were the primary receptors for \textit{active electrolocation}. 
In our model, Mormyromast channels encoded distortions caused by the agent's own EOD, yielding a \textit{self-image}, as well as distortions associated with conspecific EODs, yielding a \textit{cons-image}.
We simulated \mbox{$N_m=36$} Mormyromasts\mormVirtualNote{}, 30\% of which were concentrated near the front $60$ degrees of the agent, since the ``chin'' on real WEF is known to be a foveal region in terms of receptor density and enhanced forward-facing sensing \citep{von1999active}. 
The remaining sensors were uniformly distributed along the circumference of the agent.

\subsubsection{Ampullary sensors} 
These were modeled as \textit{passive} sensors that in real WEF sense static or low-frequency electric fields (below 40 Hz \citep{benda_physics_2020}) from nearby prey and conspecifics. 
We simulated $N_a=24$ Ampullary sensors that were distributed uniformly along the circumference of the agent.

\subsubsection{Knollenorgan sensors}
These sensors provided passive detection of conspecific EOD communication signals. 
In real WEF, they are highly sensitive to the sharp, high-frequency EOD signals emitted by conspecifics.
In our agents, these sensors were only sensitive to the electric fields generated by other agents' EODs, and were insensitive to everything else.
We simulated $N_k=12$ uniformly spaced Knollenorgan sensors per conspecific.
This allowed the agent to independently sense EODs from each conspecific within each simulation step (12 ms) without collisions in time.
Knollenorgan sensors provide waveform intensity information whose magnitude is believed to carry information about conspecific size and sex \citep{von2002imaging}.
Conspecific size metadata was therefore included in the corresponding observation channels, as described below.

\subsubsection{Sensor processing}

\paragraph{Field transduction.}
For each sensor, the electric field impinging on the sensor was converted to a raw measurement by projecting the field vector onto the sensor's surface-normal vector. 
Each sensor's normal vector was directed radially outwards from the center of the agent's circular body.
This projection captured the component of the field that the sensor was oriented to detect:
$$ s = \mathbf{E}(\mathbf{r}_s) \cdot \mathbf{n} $$
where $\mathbf{E}(\mathbf{r}_s)$ is the electric field vector at the sensor's position $\mathbf{r}_s$, and $\mathbf{n}$ is the sensor's outward-facing normal vector.

\paragraph{Baseline subtraction.}
To prevent self-EOD and cons-EOD fields from saturating sensor channels, we subtracted precomputed sensor-specific baselines from raw field readings. Baselines were computed once at simulation start and differed by sensor type.

\textit{Ampullary sensors}: We subtracted two baselines from each raw reading. First, the agent's own intrinsic dipole field was always subtracted, suppressing static self-field components. Second, when the agent emitted an EOD, we subtracted the self-EOD corollary-discharge component.

\textit{Mormyromasts}: \mormBaselineSentence{}

\textit{Knollenorgans}: These channels were tuned to conspecific EOD transients and did not require explicit baseline subtraction.

\paragraph{Scaling and normalization.}
After baseline subtraction, each sensor reading $s$ was mapped to a signed, unitless value in $[-1,1]$ via sign-preserving, magnitude-only normalization over that sensor's dynamic range $[|E_{\min}|, |E_{\max}|]$.

Let $\operatorname{sgn}(s)$ denote the sign of $s$, and define the clipped magnitude
$$\tilde{m} \;=\; \mathrm{clip}\!\left(|s|,\;E_{\min},\;E_{\max}\right).$$
We then computed the normalized log-magnitude
\[
\hat{m} \;=\;
\dfrac{\log_{10}\!\big(\max(\tilde{m},\,\varepsilon)\big)-\log_{10}(E_{\min})}
{\log_{10}(E_{\max})-\log_{10}(E_{\min})},
\]
where $\varepsilon \le E_{\min}$ is a small constant for numerical stability (default $\varepsilon=10^{-25}$). The final normalized reading is
\[
\hat{s} \;=\; \operatorname{sgn}(s)\,\hat{m}\;\in\;[-1,1].
\]
This preserves field polarity while compressing dynamic range for stable policy optimization.

\subsubsection{Sensor calibration}

We calibrated the intrinsic electric sources of agents and prey, and EOD monopole source strengths, to ensure that each class of electrosensor could detect objects within biologically plausible spatial ranges using commonly assumed sensitivity thresholds (Table~\ref{tab:calibrated}).

As a starting point, we used analytical expressions for dipole fields and induced dipole interactions (Appendix~\ref{supp:calibration_calculations}), 
required for each sensor type to reach its respective sensitivity threshold at a target detection distance.  
We then tuned these values empirically using simulations of one or two fish and a single food object. 

For Ampullary sensors (passive), we tuned intrinsic charges (using analytic expressions, see Appendix \ref{supp:calibration_calculations}) on food and prey to detect low-frequency electric fields under 40 Hz from prey up to 4 cm away and from conspecifics up to 8 cm away. 

For Mormyromasts, calibration was baseline- and dynamic-range based.
\textbf{Self-images:} we first estimated the self-EOD operating baseline at each sensor, then set dynamic ranges to detect baseline perturbations from prey up to 5 cm and conspecifics up to 10 cm (lower bound $\sim \text{baseline} \times 10^{-6}$; upper bound $\sim \text{baseline} \times 0.25$).
\mormConsCalibSentence{}

Knollenorgans, which sense high-frequency EOD pulses from conspecifics, were tuned to detect signals up to 100 cm and encoded only the sign of the local electric field.
We denote the resulting Mormyromast and Knollenorgan conspecific-sensing ranges by $R_M$ ($\approx 10$ cm) and $R_K$ ($\approx 100$ cm), respectively; these define the proximity conditions used in the decoding analyses (Results).

These calibration choices preserved the intended functional specialization of the three modeled receptor classes: Ampullary sensors provided broad passive sensitivity at short range, Mormyromasts provided spatially resolved active and collective electrolocation over shorter ranges, and Knollenorgans provided long-range directional cues restricted to conspecific EODs.

\subsection{Agent observations and actions}
The observation vector at each timestep comprised: (i) \mormObsEntry{}, (ii) $N_a = 24$ Ampullary readings, (iii) $N_k \times (N_{agents}-1) = 12 \times 3$ Knollenorgan readings plus a scalar encoding the emitting agent's size for each conspecific, (iv) the agent's last action vector, (v) egocentric displacement vectors, and (vi) the agent's own size (scalar context input).
This gives a fixed-length observation vector for the default $N=4$ group; when evaluating with fewer agents the unused conspecific slots in the Knollenorgan block are zero-padded.
Agents moved according to first-order command kinematics; movement outputs directly set bounded translational and rotational velocity commands.
The bounds on translational and rotational speed were derived from empirical tracking data on real weakly electric fish and were scaled by the agent's size parameter.
A bite event, available when the thresholded bite command was active and a conspecific was within the fixed biting radius, yielded a close-range aggression event with reward consequences.
Food consumption was handled separately and did not result from biting another agent.

\subsection{Training}

\paragraph{Algorithm.}
Agents were trained jointly using a shared-parameter variant of Proximal Policy Optimization (PPO) \citep{schulman2017proximal,yu2022surprising}, with Generalized Advantage Estimation (GAE) for variance reduction \citep{ni2021recurrent}.
Training was conducted over 20 independent random seeds.
After training, seeds were ranked using the ethologically motivated selection criterion described below, rather than by training reward alone.
Hyperparameter details are given in Supplementary Section~\ref{supp:training}.

\paragraph{Reward structure.}
Each agent received an individual reward signal comprising: 
(1) a positive reward for consuming food; 
(2) a penalty for being bitten, scaled by the size difference between biter and victim in heterogeneous groups;
(3) a small cost for successful biting; 
(4) a collision penalty; 
(5) a shaping term proportional to the decrease in distance to the nearest food item, normalized by arena perimeter; and
(6) an effort-over-threshold penalty for large movement or turn commands. 
No explicit reward for social communication or coordination was provided; any emergent signaling arose solely from the shared environment and individual fitness incentives.
The proximity-shaping term provided a dense learning signal for approach behavior under what would otherwise be a sparse-reward foraging task, and was included for sample-efficiency in line with standard RL practice on long-horizon navigation problems.
Reward coefficient values are given in Supplementary Section~\ref{supp:reward}.

\paragraph{Homing task.}
For the homing validation (Fig.~\ref{fig:homing}), we separately trained agents in a minimal two-agent setup whose reward comprised a proximity reward for reaching the partner, a small distance-decrease shaping term, and a small time penalty, with no foraging or aggression terms.
At evaluation, the target agent was fixed and always emitting while the homing agent approached from random initial positions.
The target agent was also fixed during training, matching the fixed-target evaluation assay.
For comparison in Fig.~\ref{fig:homing}, the ray-casting baseline was an alternative non-electric perception model in which radial rays returned object identity and distance-to-intersection cues for food, walls, or other agents.
Unlike the electrosensing model, this baseline did not compute electric fields or induced dipoles, and therefore provided direct geometric line-of-sight information rather than physics-based electroreceptor measurements.
For linear decoding analyses of homing performance (Fig.~\ref{fig:rnn}D), the Knollenorgan error angle $\Theta^K_\mathrm{error}$ was defined as the signed angular difference between the agent's heading and the direction of the electric field produced by the target agent at the homing agent's sensor array.

\paragraph{Neural network architecture.}
Each agent had independent actor and critic networks, each comprising an MLP encoder followed by a single-layer GRU \citep{cho2014learning} ($D_H = 512$), whose hidden state was carried across timesteps.
The actor network's hidden state drove a mixed continuous-discrete action distribution (translation, rotation, EOD emission probability, bite probability); the critic network's hidden state produced a scalar state-value estimate.
Full architecture details are given in Supplementary Section~\ref{supp:architecture}.

\paragraph{Seed selection.}
We trained 20 independent random seeds using the same architecture, environment, and reward function for 5M environment steps each, yielding variably performing agents at convergence of the high-dimensional stochastic nonconvex optimization \citep{huang2024measuring}. 
After training, seeds were evaluated on a fixed assay of test rollouts that was not used for parameter updates. 
We selected a representative seed using a pre-specified ethologically motivated ranking procedure designed to identify policies that both performed well and expressed experimentally observed size-dependent resource asymmetries.

For each seed and rollout, we computed three summary metrics. 
First, we measured total food consumed by all agents, corresponding to social foraging performance and individual-fitness maximization. 
Second, we fitted a linear model relating food consumed by each agent to that agent’s size,
\[
\mathrm{food\_eaten}_i
= \beta_0 + \beta_1 \mathrm{agent\_size}_i + \epsilon_i,
\]
and summarized the seed by the median slope ($\beta_1$) across rollouts. 
This metric measures alignment with the experimental observation that larger agents obtain more food. 
Third, we computed the median ($R^2$) of this linear fit across rollouts, measuring the consistency of the size–food relationship.

To exclude degenerate policies in which one or more agents failed to forage, we filtered out seeds whose median intercept from the food-eaten versus agent-size fit was non-positive. 
Remaining seeds were ranked using the three evaluation metrics: total food consumed, median size–food slope, and median fit ($R^2$). 
As a consequence, the size–food relationship summarized in Fig.~\ref{fig:behavior}M is best seen as a quantitative confirmation, not an independent finding. 
The combined top-ranked seed was used for the detailed behavior, perturbation, and recurrent-dynamics analyses reported in the Results.

\subsection{Evaluation and analysis}

\subsubsection{Evaluation protocol}
Post-training evaluations were performed with fixed network weights.
Evaluation settings were assay specific. The main 4-agent patchy and uniform foraging assays used 30 episodes in 70 cm $\times$ 70 cm arenas. The homing assay used 100 episodes with pre-specified initial positions in a 70 cm $\times$ 70 cm arena, with the homing agent initialized at least 20~cm from the target and both agents at least 5~cm from the walls. Two-agent square assays used 70 cm $\times$ 70 cm arenas, the two-agent wide assay used a 160 cm $\times$ 40 cm arena, and one-patch assays used a single centrally placed replenishing food patch. Initial positions, orientations, and size settings were randomized or fixed according to the protocol for each assay.
Specific arena and food configurations for each analysis are described in the corresponding results subsection.

\subsubsection{Inter-discharge interval extraction and fitting}
The interdischarge intervals from real fish were derived from published dyadic behavior \citep{chrtkova2025unsupervised}. Published dyadic position labels were converted to interdischarge intervals using the 50 kHz sampling rate of the recordings.

IDIs from artificial agents were derived from the two-agent wide assay used for Fig.~\ref{fig:behavior}C--J, with two agents in a 160 cm $\times$ 40 cm arena and 100 evaluation episodes.

Powerlaw fits were calculated via the Python powerlaw package \citep{alstott2014powerlaw}.

\subsubsection{EOD rate and peri-event analyses}
EOD rate was computed per timestep and binned by context variable (distance to nearest agent, food item, or wall) using 15 uniform bins spanning the relevant distance range; each bin reports mean $\pm$ SEM across observations.
For peri-event analyses, EOD rate was aligned to eating and biting events; a symmetric window of $\pm 100$ timesteps ($\pm 1200$~ms at 83~Hz) around each event was extracted and averaged across events with a minimum of 3 qualifying events per agent.


\subsection{Ablation and intervention experiments}
\paragraph{Food abundance experiments.}
Agents were evaluated in 70 $\times$ 70 cm arenas with patchy and uniform food distributions under the food-density settings used for each assay (Fig.~\ref{fig:ablations}).  
Food-scarcity sweeps included reduced patchy-food multipliers of $0.5\times$ and $0.25\times$ for the active sensor combinations used in the intervention analyses.

\paragraph{$N$-patch experiments.}
In a second food-manipulation assay, agents were evaluated in 70 $\times$ 70 cm arenas containing a fixed number of patches, $N \in \{1,2,3,4,5\}$.
We used two sweeps: an iso-food sweep, where the per-patch food multiplier was scaled so that total food remained approximately fixed across $N$, and an unconstrained sweep, where each patch used the standard food multiplier so that total available food increased with $N$.

\paragraph{Sensor ablations.}
Sensor ablation experiments examined the effect of changing access to Mormyromast, Ampullary, and Knollenorgan channels.
The intervention set includes single-class removals from the full model, mixed two-sensor controls, Knollenorgan-only and all-sensor-off controls.
These evaluations were conducted in 70 $\times$ 70 cm patchy arenas with an intermediate amount of food unless otherwise specified for a given assay.
The reduced arena size ensured that agents remained within Knollenorgan range of one another.
A parallel one-fish assay (single agent in isolation, no conspecifics present; Fig.~\ref{fig:ablations}A) was run alongside the four-agent ablations (Fig.~\ref{fig:ablations}B) so that sensor effects on individual foraging could be separated from those requiring social interaction.

\paragraph{EOD muting.}
EOD muting experiments assessed the role of active electrocommunication in group foraging and aggression.
One of the four agents had its EOD emission silenced for the duration of the evaluation; all other policy components remained intact, so the muted agent could still sense and move but emitted no EODs, and its conspecifics received no EOD-related input from it.
The intact group (all four agents emitting) served as the control condition.
Experiments were conducted in the standard 70~$\times$~70~cm patchy arenas used for the sensor ablation assays.

\paragraph{Collective sensing ablations.}
Collective sensing ablations isolated the contributions of self-generated and conspecific-generated EOD images in the Mormyromast channel.
Weakly electric fish have the capacity to sense both from self-generated EODs (self-EODs) as well as from conspecific EODs (cons-EODs) \citep{pedrajaCollectiveSensingElectric2024}.
We manipulated the sensing capabilities of the mormyromasts to create three conditions: 
(1) control, with both self- and cons-EOD sensing enabled; 
(2) self-EOD only, where agents cannot benefit from other agents' EODs; and 
(3) cons-EOD only, where agents only received feedback through the mormyromast channel from EODs generated by other agents.
These manipulations did not affect the ampullary channel.
These experiments were conducted in a very small 40 $\times$ 40 cm arena to ensure close proximity, as the cons-EOD benefits were expected only when agents were close to one another.
Food was uniformly distributed at intermediate density.
Each condition received 30 evaluation runs.

\subsection{Two-agent experiments}
\paragraph{Two-agent foraging.}
Two-agent experiments allowed closer examination of how EOD modulation and relative-size asymmetries shape pairwise competition.
We evaluated two agents in a 70 $\times$ 70 cm square arena with intermediate uniform food abundance for pairwise interaction and recurrent-state analyses. Agents were initialized with random positions, orientations, and size settings.
We also evaluated agents in a \textit{wide} 160 $\times$ 40 cm arena to quantify long-range EOD modulation, IDI statistics, and attraction or repulsion between two agents.
Agents were initialized in a 160 $\times$ 40 cm arena, one on the left at (20, 20) and the other on the right side of the arena at (140, 20).

\paragraph{Bot (random-walker) control.}
To isolate social cues from food-associated gradients, in a subset of two-agent assays we replaced the resident agent with a rule-based bot confined to the food patch.
The bot performed a random walk within the patch region and did not respond to its partner or to any sensory input.
We ran conditions in which the bot was either stationary (\emph{frozen}) or mobile, and in which it emitted EODs according to a simple stochastic schedule at a controlled rate (including a non-emitting, zero-rate condition).
This provided a partner that supplied food-associated and EOD cues but no adaptive social behavior, and whose emission timing differed in temporal structure from that of trained agents.

\paragraph{One-patch dyadic assay.}
To assess resource-access competition between two agents differing in body size, we evaluated agents in a one-patch arena with a single centrally placed replenishing food patch (150 cm $\times$ 150 cm; see Supplementary Section~\ref{supp:arena}).
Agent A (resident) was initialized on the patch; agent B (intruder) was initialized at a random position within Knollenorgan detection range (${\le}100$ cm from A).
We tested three relative size conditions: $A < B$ (intruder larger), $A = B$ (size-matched), and $A > B$ (resident larger), plus bot-control variants in which the resident was replaced by the random-walking bot described above.

We defined pairwise interactions between agents $A$ and $B$ using their allocentric positions $\mathbf{r}_A(t),\mathbf{r}_B(t)$ and orientations $\theta_A(t),\theta_B(t)$ at each timestep $t$.

\paragraph{Interaction bouts.}
An interaction bout between two agents was defined as a maximal contiguous segment of time during which the agents remained within close spatial proximity. Bout start and end times were determined by the first and last timesteps of such proximity.

\paragraph{Distance thresholds.}
During each bout we tracked the inter-agent distance $d_{AB}(t)$. Two distance thresholds were used: a \textbf{contact threshold} ($d_{\mathrm{contact}}=2$\,cm), indicating physical contact if $d_{AB}(t)\le d_{\mathrm{contact}}$ at any timestep, and a \textbf{reduced interaction threshold} ($d_{\mathrm{reduce}}=5$\,cm), used when computing distance-restricted interaction statistics.

\paragraph{Geometric interaction labels.}
At each timestep we classified the spatial relationship between the two agents based on the alignment of each agent's heading relative to the bearing toward the other agent.

Let $\phi_{A\rightarrow B}(t)$ denote the bearing from $A$ to $B$, and $\phi_{B\rightarrow A}(t)$ the bearing from $B$ to $A$. We computed the angular difference between each agent's orientation and the bearing to its partner:
\[
\Delta_A(t) = \theta_A(t) - \phi_{A\rightarrow B}(t), \qquad
\Delta_B(t) = \theta_B(t) - \phi_{B\rightarrow A}(t).
\]

Using angular thresholds $\theta_{\mathrm{close}}=\pi/4$ and $\theta_{\mathrm{opp}}=3\pi/4$, interaction states were defined as:
\textbf{Confronting} occurred when both agents faced toward each other ($|\Delta_A|<\theta_{\mathrm{close}}$ and $|\Delta_B|<\theta_{\mathrm{close}}$). 
\textbf{Chasing} occurred when agent $A$ faced toward $B$ while $B$ faced away ($|\Delta_A|<\theta_{\mathrm{close}}$ and $|\Delta_B|>\theta_{\mathrm{opp}}$). 
\textbf{Fleeing} occurred when agent $B$ faced toward $A$ while $A$ faced away ($|\Delta_B|<\theta_{\mathrm{close}}$ and $|\Delta_A|>\theta_{\mathrm{opp}}$). 
\textbf{Dispersing} occurred when both agents faced away from each other ($|\Delta_A|>\theta_{\mathrm{opp}}$ and $|\Delta_B|>\theta_{\mathrm{opp}}$). 
All other relative orientations were classified as \textbf{unaligned}.

Thus, in a \textit{chasing} configuration agent $A$ is the chaser and $B$ the chased, whereas in a \textit{fleeing} configuration the roles are reversed.

\paragraph{Bout-level summaries and biting.}
Each interaction bout was assigned a dominant interaction class defined as the most frequent timestep label during the bout. Aggressive encounters were defined as bouts in which either agent performed a bite action or was bitten by its partner at any point during the interaction.

\subsection{Neural analyses}

\subsubsection{Effective dimensionality}
We quantified the effective rank of RNN population activity using the participation ratio of PCA explained variance ratios:
\[
D_\mathrm{eff} = \frac{1}{\displaystyle\sum_i \tilde{\lambda}_i^2}
\]
where $\tilde{\lambda}_i$ is the fraction of variance explained by the $i$-th principal component (eigenvalues normalized to sum to 1).
$D_\mathrm{eff}$ was computed per episode by running PCA on the $T \times D_H$ hidden-state matrix ($T$ timesteps, $D_H=512$ units), then averaged across episodes and environments.

For group-size comparisons (Fig.~\ref{fig:rnn}B), the same trained policy was evaluated in separate rollouts with $N \in \{1, 2, 3, 4\}$ agents in 70~cm $\times$ 70~cm patchy arenas; $D_\mathrm{eff}$ was computed per episode and averaged across episodes within each condition.
To verify that group-size differences in $D_\mathrm{eff}$ were not an artifact of the number of simultaneously recorded units, we performed unit subsampling (Fig.~\ref{fig:rnn}C): for each group size we repeatedly ($n=30$ per fraction) drew random subsets of hidden units at fractions $\{0.05, 0.15, 0.25, 0.5, 0.75, 1.0\}$ and recomputed $D_\mathrm{eff}$ from the subsampled activity.

\subsubsection{Feature extraction}
We extracted a broad panel of hand-curated features spanning
egocentric geometry (orientation, turn angle, displacement),
electrosensory inputs (Knollenorgan, Mormyromast, and Ampullary field angles and magnitudes),
distances to agents, food, and walls,
temporal recency of key events (time since last EOD, feeding, or encounter),
boolean task features (nearby agents, bites, emissions, eating), and
local counts of food or agents.
Together, these circular, scalar, boolean, probability, and count features capture the sensory, spatial, and behavioral context available to the agent at each timestep.
The full set of features is listed in Table~\ref{tab:features}.

\subsubsection{Decoding analyses}
We quantified how individual task and behavioral variables could be linearly decoded from the agents' recurrent states.
For each target, we fit ordinary least-squares linear models from recurrent hidden state to the target value and evaluated generalization on held-out episodes using grouped cross-validation.
Scalar, boolean, probability, and count targets were treated as real-valued regression targets and scored with held-out $R^2$.
Circular targets were decoded by fitting parallel linear models to $\cos \phi$ and $\sin \phi$ and averaging the held-out $R^2$ across the two components.
For each cross-validation fold, we subtracted the performance of a mean-prediction baseline fit on the training episodes, yielding a baseline-normalized test score.
Reported error bars for decoding panels show the standard error across cross-validation folds.


\subsubsection{Partial least squares correlation (PLSC)}
We used partial least squares correlation (PLSC) to quantify shared population structure in the activity of pairs of RNNs. 
For every episode with at least one second of pairwise interaction (agents within the Mormyromast conspecific-sensing range $R_M$), we concatenated agent hidden states aligned by time and made balanced ``in-range'' and ``out-of-range'' datasets.
After z-scoring hidden-state features, we fitted PLSC to find shared axes of variation and then computed the Pearson correlation between the paired canonical variates. 
Significance was assessed against a null distribution constructed by circularly shifting one agent's hidden-state sequence in time ($n{=}30$ shifts), preserving temporal autocorrelation while disrupting alignment between agents.
A component was counted as significant only when both its singular value and projected correlation exceeded the $97.5^{th}$ percentile of the corresponding null distributions (one-sided test).
The number of significant dimensions was counted contiguously from the first component until the first non-significant component.
Additional details on PLSC are in Supplementary Section~\ref{supp:plsc_math}.

\subsubsection{Power spectral density}
We quantified wideband power in recurrent hidden-state activity for each episode, environment, and agent.
Hidden-state time series were transformed with a short-time spectrogram using 20-timestep windows with 10-timestep overlap, and power was averaged across recurrent units, time windows, and frequencies to obtain one mean wideband-power value per agent episode.
These episode-level power values were associated with timestep-level social labels to compare activity during periods with and without a nearby conspecific, and with agent-level size status in two-agent assays.
For size-status comparisons, the larger member of a pair was labeled as the larger agent and the smaller member as the smaller agent.

\subsection{Statistical analyses and annotation}
Statistical summaries were computed at the level appropriate to each assay: evaluation episodes, agent episodes, interaction bouts, agent pairs, events, or cross-validation folds.
Food consumption inequality within groups was quantified using the Theil T index
\[
T = \frac{1}{N}\sum_{i=1}^{N} \frac{x_i}{\mu}\ln\!\frac{x_i}{\mu},
\]
where $x_i$ is food consumed by agent $i$ and $\mu$ is the group mean; $T=0$ indicates perfect equality \citep{conceiccao2000young}.
In biting analyses, the win ratio for each agent is the fraction of biting events involving that agent in which it was the biter:
$\text{win ratio} = n_{\text{given}} / (n_{\text{given}} + n_{\text{received}})$.
For binned EOD-distance and peri-event EOD analyses, curves show mean $\pm$ SEM (standard error of the mean) across observations or events entering each bin or time offset.
For all box-and-whisker plots, where $\leq 50$ observations are being plotted, we have superimposed the jittered scatterplot on the plot.
For intervention panels comparing each manipulation to a control condition, significance annotations use Dunnett one-vs-control tests.
For two-group unpaired comparisons, we used two-sided Mann-Whitney tests.
For paired role comparisons within the same interaction, we used two-sided Wilcoxon signed-rank tests.
Linear decoding analyses report held-out, baseline-normalized $R^2$ with SEM across grouped cross-validation folds.
PLSC significance was assessed with the circular-shift null described above.
Unless otherwise specified, tests were treated as panel-level exploratory comparisons and no additional correction was applied across figure panels.
Statistical annotations use the following convention:
\begin{center}
\begin{tabular}{r l}
\toprule
Symbol & Significance range \\
\midrule
ns   & $5.00\times10^{-2} < p \le 1.00$ \\
$*$    & $1.00\times10^{-2} < p \le 5.00\times10^{-2}$ \\
$**$   & $1.00\times10^{-3} < p \le 1.00\times10^{-2}$ \\
$*{*}*$  & $1.00\times10^{-4} < p \le 1.00\times10^{-3}$ \\
$*{*}{*}*$ & $p \le 1.00\times10^{-4}$ \\
\bottomrule
\end{tabular}
\end{center}




\section*{Data and Code Availability}
Training, simulation, and analysis code, as well as examples of trained policies, are available at 
\newline 
\url{https://github.com/KempnerInstitute/wef} 

\section*{Ethics and Reproducibility}
All experiments are in silico. 
Our framework reduces invasive experimentation on live animals by enabling in silico hypothesis generation and prioritization.  
All experimental data used in this paper were collected by collaborators for previous neuroscientific studies of weakly electric fish.

\section*{Author Contributions (CRediT Taxonomy)}
S.H.S. and S.J.Y.: conceptualization, methodology, software, formal analysis, visualization, writing -- original draft. 
Z.L., A.W., F.P., and D.T.: methodology, software. 
P.S. and N.S.: writing -- review and editing. 
S.H.S., S.J.Y., N.B.S., and K.R.: conceptualization. 
N.B.S. and K.R.: supervision. 
K.R.: funding acquisition. 
All authors reviewed and approved the final manuscript.

\section*{Acknowledgements}
We thank Eugene Vinitsky, Yohan John, Ann Huang, Gizem Ozdil, Nathan Wu, Roy Harpaz, Ella Batty, Kianté Brantley, Na (Lina) Li, Sam Gershman, Yuyang Zhang, Thomas Fel, Daphne Cornelisse, and members of the Rajan, Brantley and Gershman labs for helpful discussions.
Funded by NIH (RF1DA056403), 
James S. McDonnell Foundation (220020466), 
Simons Foundation (Pilot Extension-00003332-02), 
McKnight Endowment Fund, 
CIFAR Azrieli Global Scholar Program, NSF (2046583), 
Harvard Medical School Dean's Innovation Award, 
Harvard Medical School Neurobiology Lefler Small Grant Award,
Alice and Joseph Brooks Fund Postdoctoral Fellowship (S.H.S.),
Shanahan Family Foundation Fellowship at the Interface of Data and Neuroscience at the Allen Institute and University of Washington, supported in part by the Allen Institute (D.T.).

\newpage
{\small\bibliography{fish}}

\clearpage
\onecolumn
\appendix
\section*{Supplementary Information}

\subsection{Related Work}
\label{supp:related_work}
Our work builds upon research spanning neuroethology, collective behavior, and artificial intelligence.

Weakly electric fish have been a key model system for understanding how animals use self-generated signals to sense their surroundings and communicate.
Pulse-type weakly electric fish (\textit{Mormyridae}) emit brief electric pulses known as electric organ discharges (EODs), whose timing and structure can be modulated depending on behavioral context \citep{caputiElectricOrganDischarge1999,carlsonElectricSignalingBehavior2002,von1999active,kramer_communication_1994}.
Their electrosensory system includes three major receptor classes: Mormyromasts for active electrolocation, Knollenorgans for passive detection of conspecific EODs, and Ampullary organs for low-frequency passive sensing.
Together, these systems allow these nocturnal fish to sense their environment in the turbid, visually challenging conditions that characterize their natural habitat \citep{benda_physics_2020}.
Work on social and agonistic behavior has shown that EOD patterns carry information about identity, dominance, and aggression \citep{mollerElectricOrganDischarge1989,carlsonStereotypedTemporalPatterns2004,bakerMultiplexedTemporalCoding2013a,stoddardPredationCrypsisEvolution2019}, while recent studies suggest that groups of weakly electric fish can benefit from one another's signals through collective sensing \citep{pedrajaCollectiveSensingElectric2024}.
At the physiological level, increasingly detailed biophysical and computational models are beginning to explain how electrosensory circuits process these signals \citep{sawtell2005sparks,fukutomiHistoryCorollaryDischarge2020,engelmannLinkingActiveSensing2021,chenModelingSignalBackground2005,wallachInternalModelCanceling2023,wallachMixedSelectivityCoding2022,turcu2025end}.
Our contribution is to place these biophysical foundations inside a closed-loop, multi-agent decision-making framework.

Classical agent-based models have shown that simple local rules can generate complex group-level phenomena in fish schools and bird flocks \citep{bod2018probabilistic,huang2024collective,Harpaz2021-fk}.
Foundational rule-based approaches identified alignment, repulsion, and attraction zones as sufficient ingredients for schooling and milling \citep{couzin2002collective}, while data-driven and deep-learning approaches have inferred interaction rules or predicted collective trajectories directly from data \citep{katz2011inferring,liReverseEngineeringControl2023,papaspyrosPredictingLongtermCollective}.
More recently, MARL has been used to model collective locomotion and schooling \citep{alageshan2020machine,loffler_collective_2023}.
While these models have provided important insights into collective dynamics, they typically rely on hand-designed interaction rules and simplified sensory representations.
As a result, they do not fully capture the closed-loop coupling between perception, motor control, communication, and ecological context that characterizes natural behavior.

Artificial neural network based RL agents have emerged as a tool for studying neural mechanisms underlying complex biological naturalistic behavior, both for individual behavior \citep{merel2019deep,keller_aran2025,vaxenburg2025whole,lobato2022neuromechfly,wang2024neuromechfly,singh2023emergent,haesemeyer2019convergent} and collective behaviors such as pursuit, swarming, flocking, resource partitioning, and cooperative hunting \citep{redman2026predictive,zhang2025inter,chen2016conceptual,chipade2021multiagent,tsutsui2024collaborative,goldshteinReinforcementLearningEnables2020,mcgraw2024parallel}.
These efforts illustrate how biologically grounded RL frameworks can provide mechanistic hypotheses that complement experimental data, consistent with broader calls to embrace naturalistic, less tightly constrained models in neuroscience \citep{lindsay2024grounding,hasson2020direct,nastase2020keep}.
Full observability and controllability of the artificial neural network enables mechanistic interpretation and in silico ablations \citep{li2024discovering,singh2023emergent,singh2021neuroprospecting}.

A related literature studies emergent communication among artificial agents trained on cooperative or competitive tasks.
In these settings, agents often develop signaling systems that improve coordination and task performance \citep{lazaridou_emergent_2020,wieczorek_framework_2024}, including under low-bandwidth communication constraints \citep{grupen_low-bandwidth_2020} or in socially competitive environments \citep{rachum_emergent_2024}.
However, these studies typically consider explicit, often symbolic, communication channels rather than signals grounded in biophysically constrained sensing.

\subsection{Arena and Foraging Environment}
\label{supp:arena}

The arena is a rectangular, bounded region with reflective walls.
During training, width and height are independently sampled from $[40, 300]$~cm, so the policy is exposed to a broad range of arena scales.
Evaluation assays use fixed sizes: 70~cm~$\times$~70~cm for square foraging and homing, 160~cm~$\times$~40~cm for the wide two-agent assay, and 150~cm~$\times$~150~cm for dyadic patch-access assays.

Four food-distribution classes were used in reported experiments:
\begin{itemize}
    \item \textbf{Uniform}: food pellets distributed uniformly across the full arena.
    \item \textbf{Patchy}: circular patches placed at a fixed spatial density; each patch radius is drawn from a truncated normal distribution.
    \item \textbf{$N$-Patch}: $N$ circular patches of fixed radius arranged evenly around an ellipse centered in the arena, with a random starting angle.
    \item \textbf{One-Patch}: a single circular patch at the arena center.
\end{itemize}

In all foraging analyses, food is initialized at episode start via a Poisson draw with density $\rho_0$ (pellets~cm$^{-2}$) and neither replenishes nor decays during the episode.
The one-patch dyadic assay is the exception: the patch replenishes continuously, sustaining competition across the episode.

\clearpage
\subsection{Dynamic Cons-Baseline Sensing Model}

A key challenge in modeling collective sensing is that the conspecific (cons-) EOD field at a receiver's Mormyromast sensors is itself a dynamic, large-magnitude baseline that varies with inter-fish distance and orientation.
The object-induced perturbation the receiver wants to detect (the \emph{cons-image}) is orders of magnitude smaller than this direct cons-EOD field.

To isolate the cons-image, we implement a \emph{dynamic cons-baseline} model inspired by the functional mechanism proposed by Pedraja and Sawtell \citep{pedrajaCollectiveSensingElectric2024}.
Whenever a conspecific emits an EOD, we compute a \emph{cons-baseline} at each receiver Mormyromast: the electric field that the conspecific's EOD would produce at that sensor location in the absence of any induced sources on nearby objects.
Subtracting this cons-baseline from the total field reading leaves a residual that reflects only the perturbations induced on surrounding objects, i.e. the cons-image:

$$s_{\text{cons-image},k} = s_{\text{total},k} - s_{\text{cons-baseline},k}$$

where $s_{\text{total},k}$ is the raw reading at Mormyromast $k$ and $s_{\text{cons-baseline},k}$ is the pre-subtracted direct cons-EOD contribution at that sensor, computed from the conspecific's EOD dipole field without object interactions.

The cons-baseline is recomputed at each timestep as the conspecific's position and emission state change, making it fully adaptive to inter-fish geometry.

\begin{figure}[htbp!]
\centering
\includegraphics[width=0.33\textwidth]{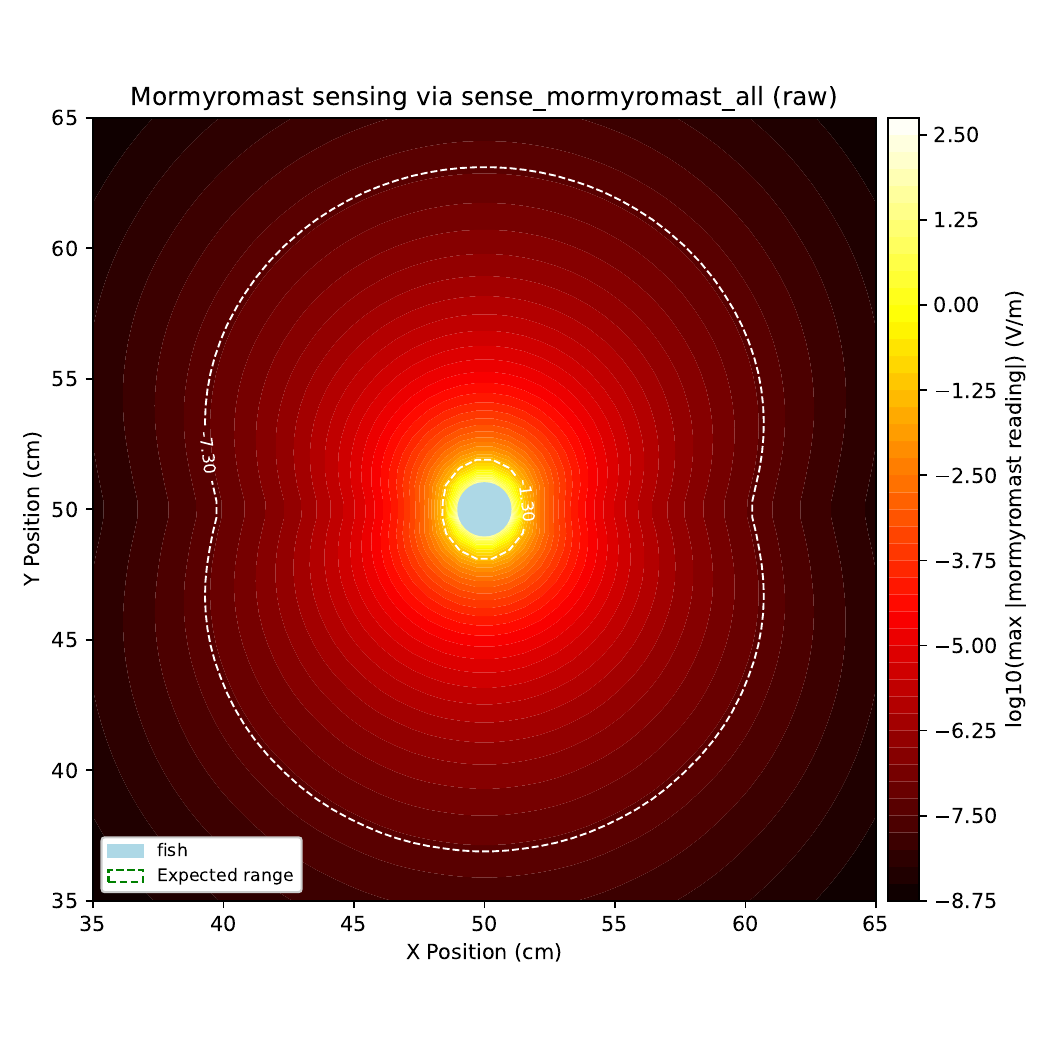}
\includegraphics[width=0.33\textwidth]{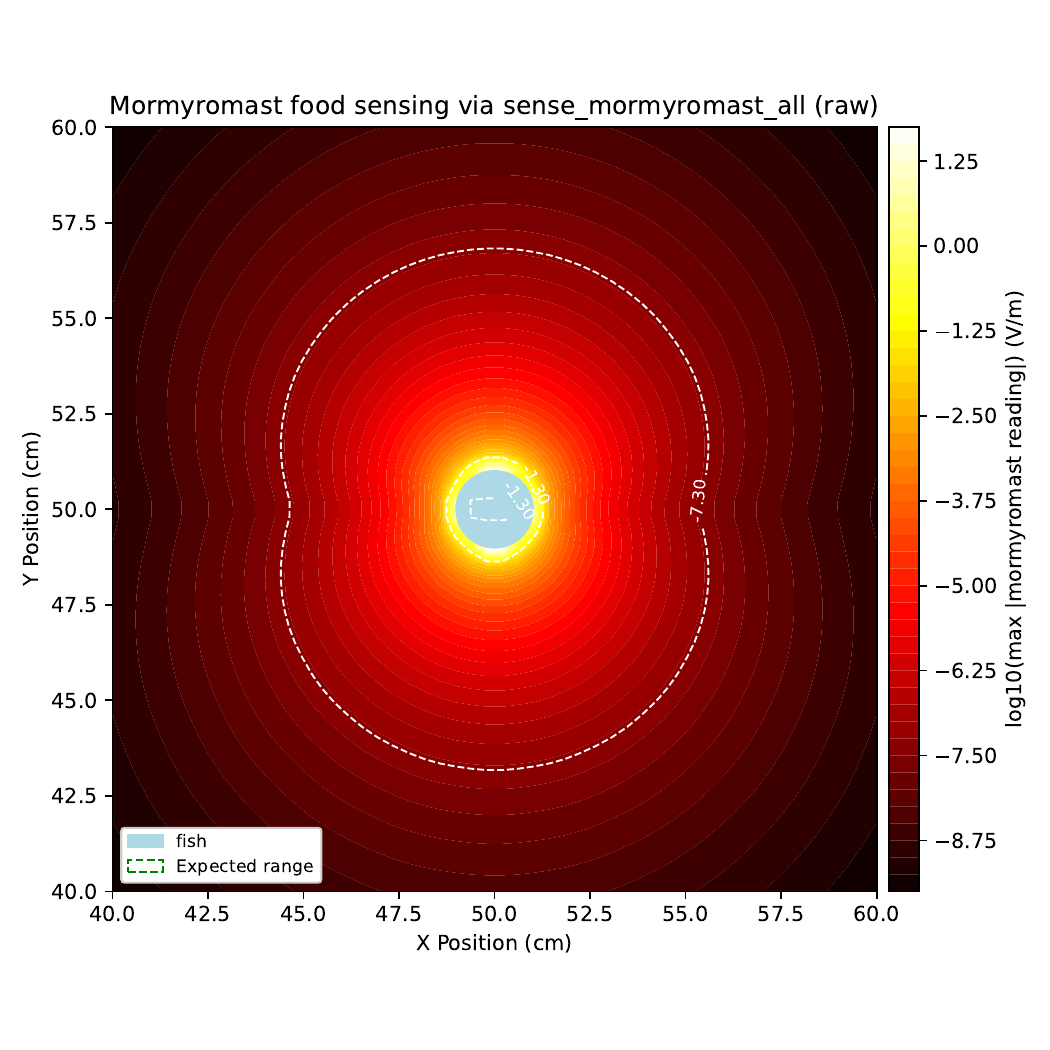} \\
\includegraphics[width=0.33\textwidth]{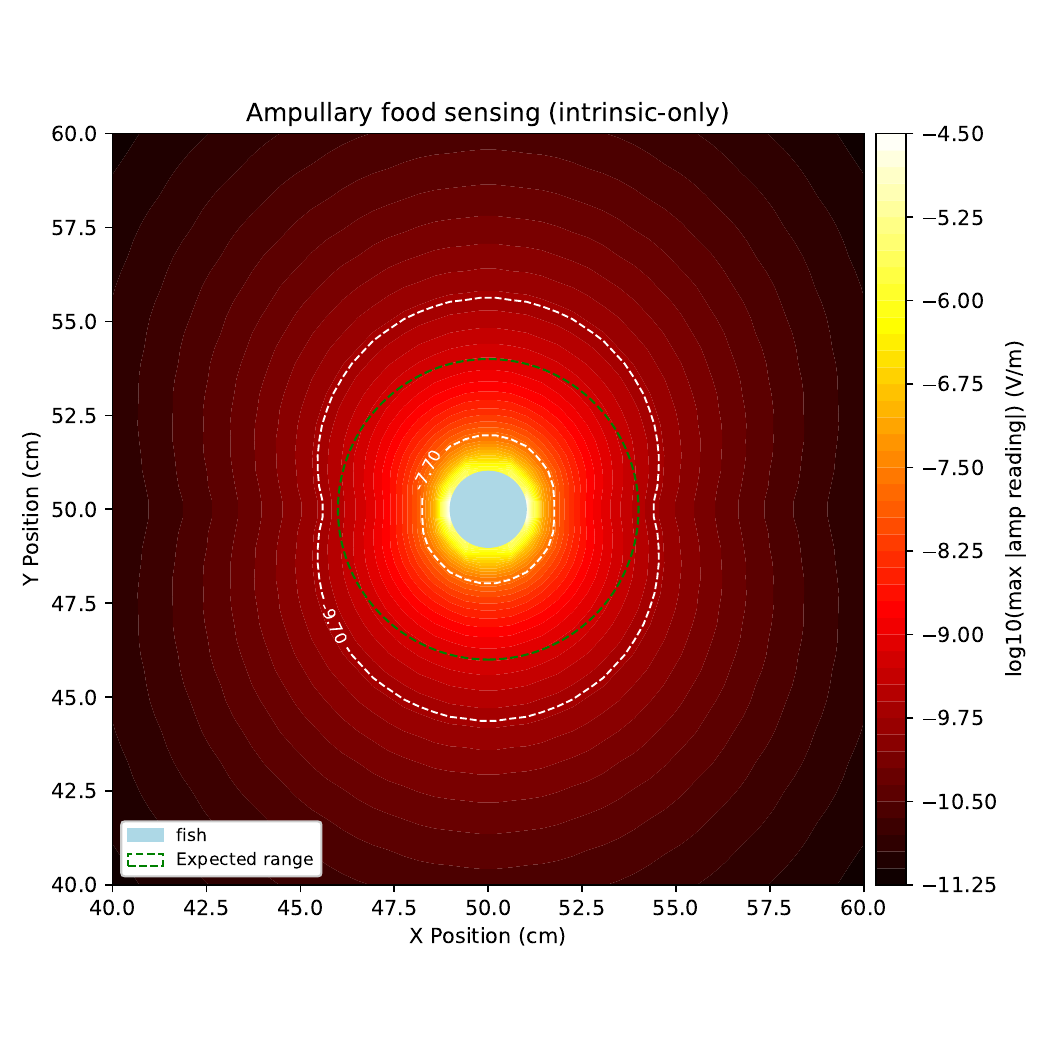} 
\includegraphics[width=0.33\textwidth]{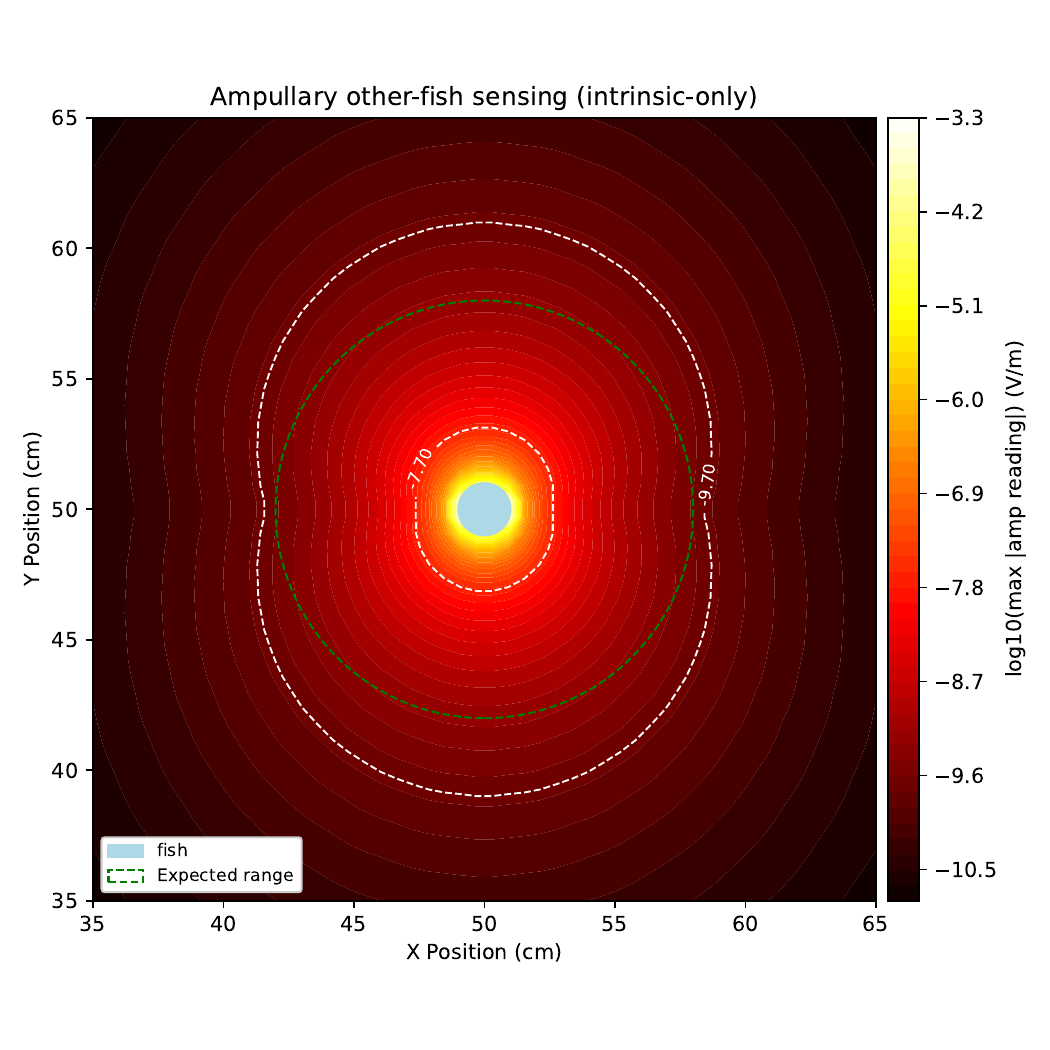} 
\includegraphics[width=0.33\textwidth]{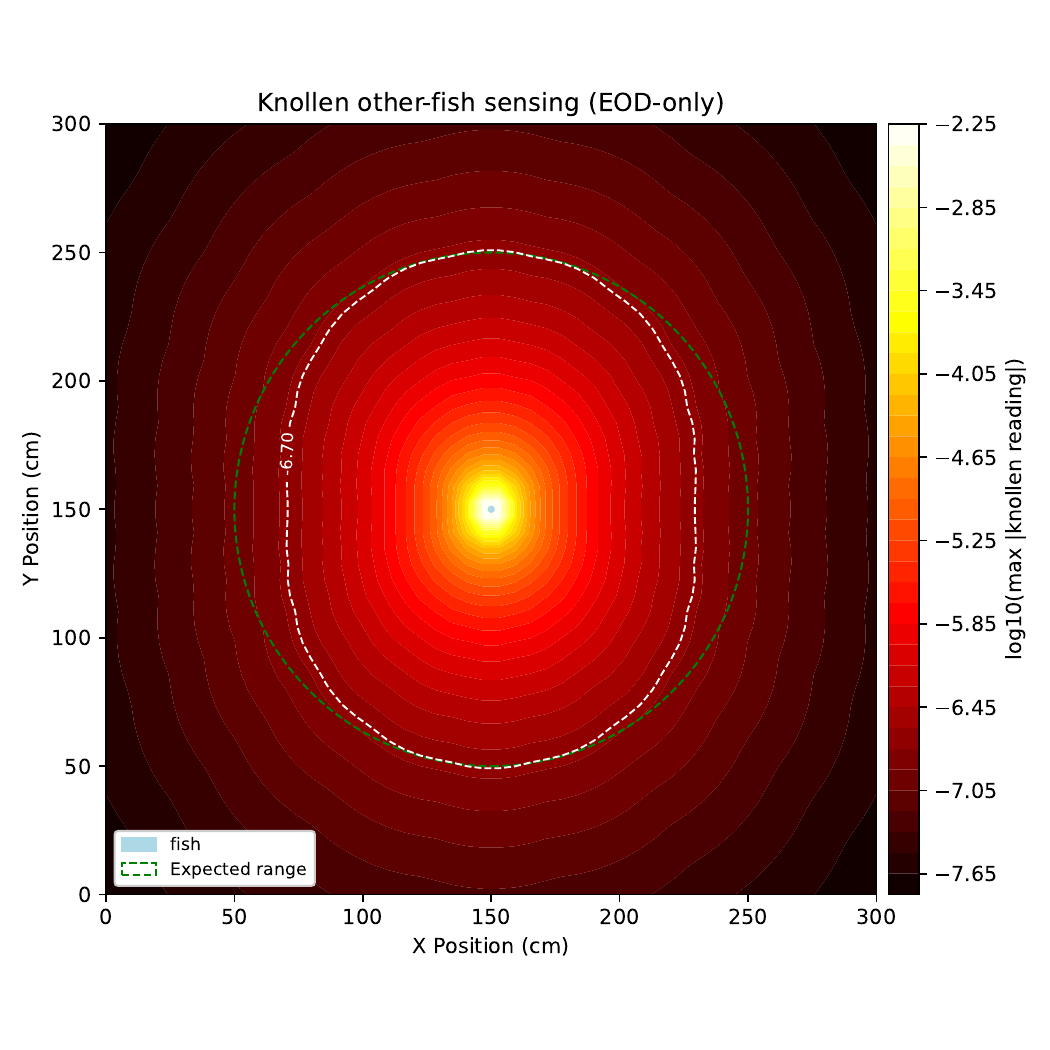} 
\caption{
Sensing ranges for different sensing modalities:
Maximum intensity of \textbf{self-images} produced by food objects [Upper Left] and conspecifics [Upper Right] when actively sensed by \textbf{Mormyromasts}.
Maximum intensity of intrinsic dipoles emanating from prey [Lower Left] and conspecifics [Lower Middle], sensed by \textbf{Ampullary} receptors.
[Lower Right] Maximum intensity of passively sensed conspecific EODs using \textbf{Knollenorgans} (that binarize the received signal).
}
\label{fig:self_and_direct}
\end{figure}
{}

\clearpage
\subsection{Multi-Agent Reinforcement Learning}

The foraging task is formulated as a decentralized partially observable Markov decision process (Dec-POMDP): at execution time each agent acts using only its local electric-sense observations with no explicit communication channel, so any emergent signaling arises solely through shared environmental consequences.
During training, the critic receives the concatenated observations of all agents (centralized training, decentralized execution; CTDE), which is the standard value-estimation strategy of MAPPO \citep{yu2022surprising}.
Each agent's Knollenorgan observation includes a scalar encoding the emitting conspecific's size, consistent with the biological finding that Knollenorgan waveform intensity carries information about conspecific size and sex \citep{von2002imaging}.

\subsubsection{Training Algorithm}
\label{supp:training}
We trained 20 independent seeds for 5M environment steps each using Proximal Policy Optimization (PPO) \citep{schulman2017proximal} extended to the recurrent, multi-agent setting \citep{yu2022surprising,ni2021recurrent}.
During training, arena width and height were each sampled independently from $[40, 300]$~cm per episode, exposing the policy to a broad range of spatial scales and food-patch densities.
Key algorithmic components:
\begin{itemize}
\item \textbf{Actor-critic architecture}: independent actor and critic networks, each with an MLP encoder, a recurrent trunk, and a linear output head (Section~\ref{supp:architecture}).
\item \textbf{Generalized Advantage Estimation (GAE)} with $\lambda = 0.95$ for variance reduction.
\item \textbf{Recurrent rollout buffer}: hidden states are carried across timesteps within an episode; truncated backpropagation through time (BPTT) is used with a fixed horizon to limit memory requirements.
\item \textbf{Clipped surrogate objective} with clip parameter $\varepsilon = 0.2$.
\item \textbf{Entropy regularization} with coefficient $c_H$ to encourage exploration.
\item \textbf{Shared policy with agent-size-specific conditioning}: all $N$ agents share a single set of weights; inter-agent heterogeneity is encoded via the scalar size context input in the observation.
\item \textbf{Seed selection}: the best-performing model across 20 seeds is selected based on mean per-episode food consumption and the slope and goodness-of-fit ($R^2$) of food eaten regressed on agent size, evaluated on held-out episodes (Fig.~\ref{fig:supp_seeds}).
\end{itemize}

Hyperparameters were taken directly from the original on-policy MARL codebase \citep{yu2022surprising}, which was tuned for continuous-action continuous-control tasks; no additional tuning was performed for this work.

Each random seed produces a slightly different solution, as expected for equilibrium-finding in non-convex multi-agent settings (Fig.~\ref{fig:supp_training}, \ref{fig:supp_seeds}).
Despite this variability, the top seeds share the same qualitative patterns: size-based dominance, structured EOD modulation, and low-dimensional RNN activity (Fig.~\ref{fig:supp_inter_seed}).
We selected seed S1 for the detailed analyses in the main text.

\subsubsection{Computational resources}
Training and evaluation were run on a workstation with a single NVIDIA RTX 5090 GPU and on a GPU cluster using equivalent hardware.
Each seed took approximately 6--9 hours to complete training and post-training evaluation.

\subsubsection{Reward Structure}
\label{supp:reward}
The individual reward at each timestep is:
\begin{equation}
r_t =
r^{\text{food}}_t
+ r^{\text{prox}}_t
+ r^{\text{bitten}}_t
+ r^{\text{bite}}_t
+ r^{\text{collision}}_t
+ r^{\text{effort}}_t
\end{equation}
where:
\begin{itemize}
\item $r^{\text{food}}_t = +c_f$ whenever the agent successfully consumes a food item within its eating zone.
\item $r^{\text{prox}}_t = c_p \cdot (d_{t-1}^{\text{food}} - d_t^{\text{food}}) / P_{\text{arena}}$: shaping reward proportional to the decrease in distance to the nearest food item, normalized by arena perimeter.
\item $r^{\text{bitten}}_t$: penalty for being bitten. In heterogeneous groups, this penalty is scaled by $(1 + s_{\text{biter}} - s_{\text{victim}})$, so being bitten by a larger agent is penalized more strongly.
\item $r^{\text{bite}}_t$: small cost applied when the agent successfully bites another agent.
\item $r^{\text{collision}}_t$: penalty applied when the agent collides with another object or boundary.
\item $r^{\text{effort}}_t$: penalty for movement or turn commands above the effort threshold.
\end{itemize}
The timestep and wall-proximity reward terms were zero in all reported experiments.
No explicit reward for EOD emission, communication, or social coordination is provided.

\subsubsection{Neural Network Architecture}
\label{supp:architecture}

\begin{itemize}
  \item \textbf{Actor network}: two-layer MLP encoder (ReLU activations, LayerNorm after each layer, hidden size 512) followed by a single-layer GRU \citep{cho2014learning} ($D_H = 512$), with a linear head producing a continuous action distribution over movement commands, an EOD logit, and a bite logit. The environment thresholds the EOD and bite channels into binary events.
  \item \textbf{Critic network}: same MLP and GRU structure, with a linear head producing a scalar value estimate.
\end{itemize}

\subsection{Analyses}

\begin{figure}[htbp]
    \centering
    \includegraphics[width=0.45\linewidth]{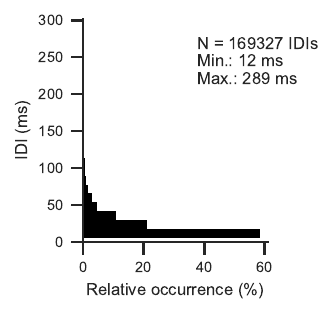}
    \includegraphics[width=0.45\linewidth]{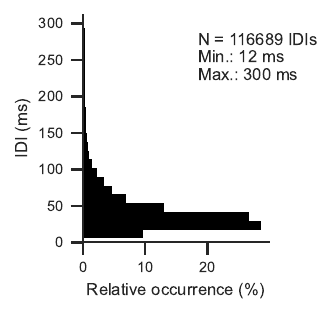}
    \caption{
    Inter-discharge intervals from two freely foraging agents [Left], and from experimental data from two fish\citep{chrtkova2025unsupervised} [Right].
    These figures are inspired by Gebhardt et al. Fig. 4C \cite{gebhardtElectricDischargePatterns2012} (and also resemble the data provided there.)
    }
    \label{fig:supp_idi_histogram}
\end{figure}

\subsubsection{Interaction definitions}
\begin{figure*}
    \centering
    \includegraphics[width=1.0\linewidth]{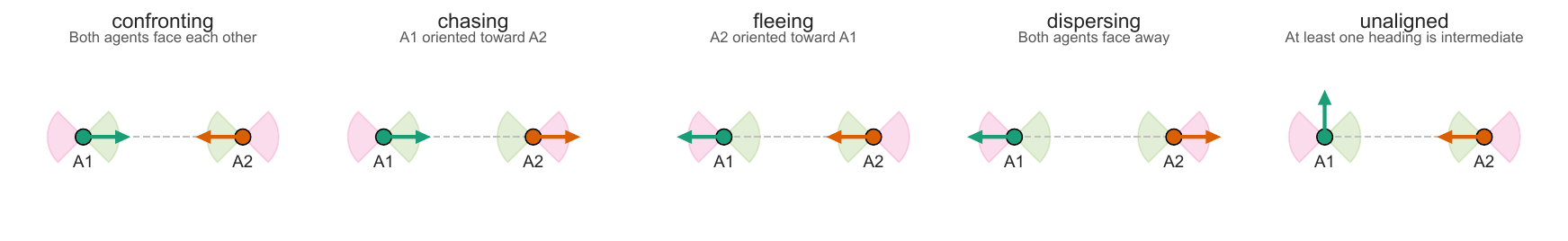}
    \caption{
    Illustrative examples of interactions between two agents (A1 and A2). 
    Green sector: $|\mathrm{rel\ angle}| < \pi/4$ (toward other).   
    Pink sector: $|\mathrm{rel\ angle}| > 3\pi/4$ (away from other).
    }
    \label{fig:supp_interactions}
\end{figure*}

\begin{figure*}[!tbp]
\centering
\panel[0.74\textwidth]{A}{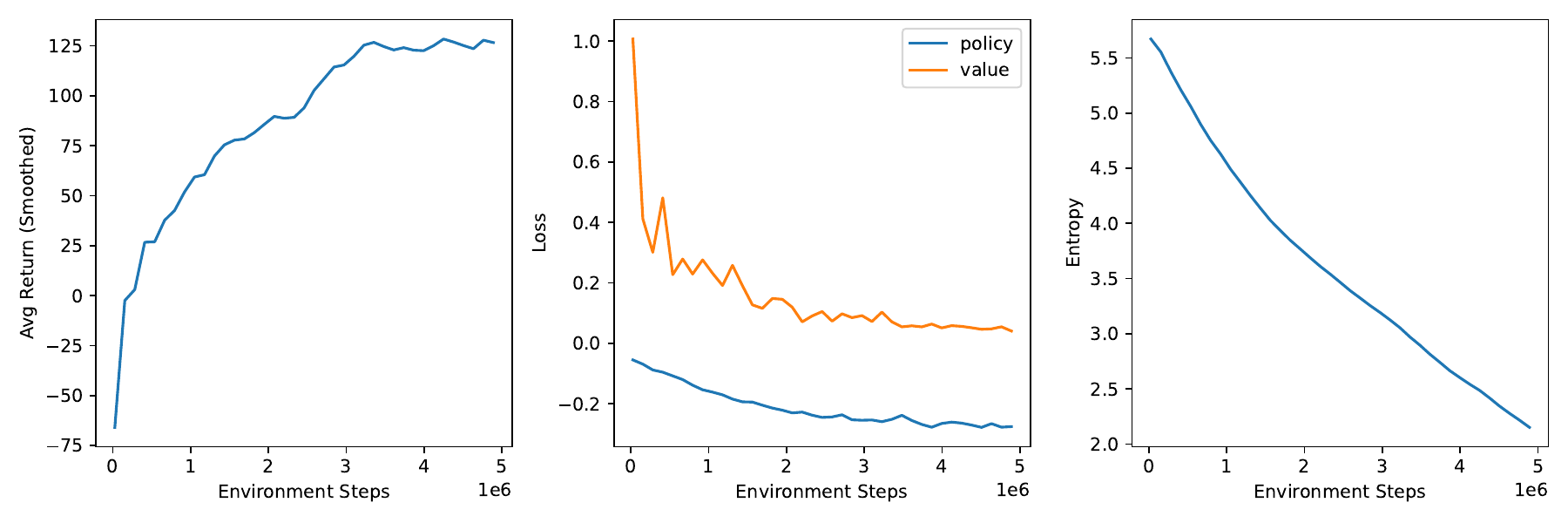}
\panel[0.25\textwidth]{B}{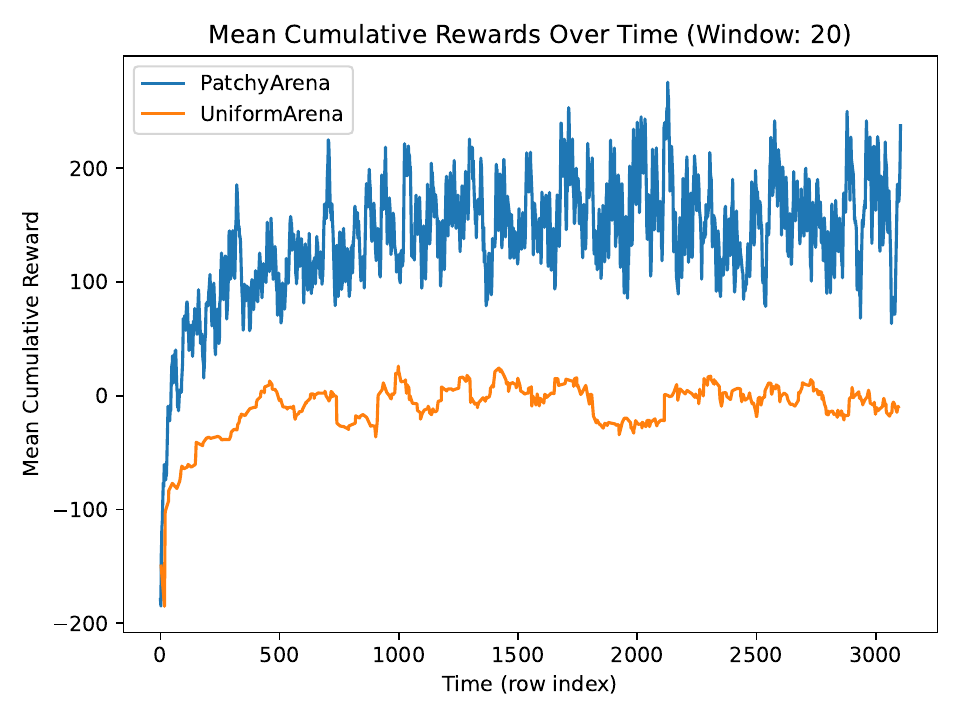}
\caption[]{
\textbf{Training diagnostics for the selected seed (S1).}
\textbf{(A)} Smoothed average episode return (left), policy and value loss (centre), and action entropy (right) over 5M environment steps, showing stable convergence.
\textbf{(B)} Mean cumulative reward (moving window of 20 updates) separated by arena type sampled during training (patchy vs.\ uniform).
}
\label{fig:supp_training}
\end{figure*}

\begin{figure*}[!tbp]
\centering
\panel[0.55\textwidth]{A}{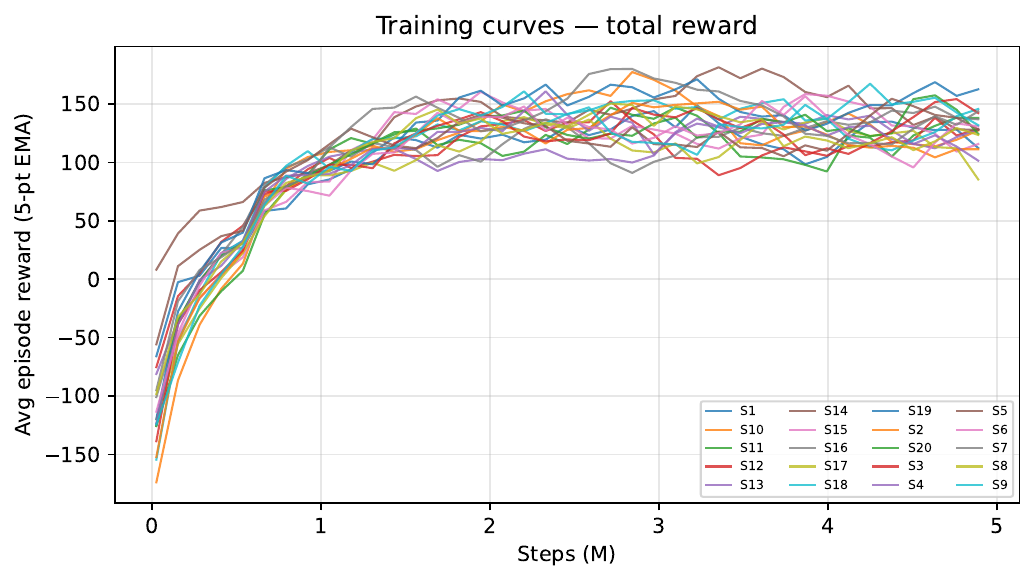}
\panel[0.25\textwidth]{B}{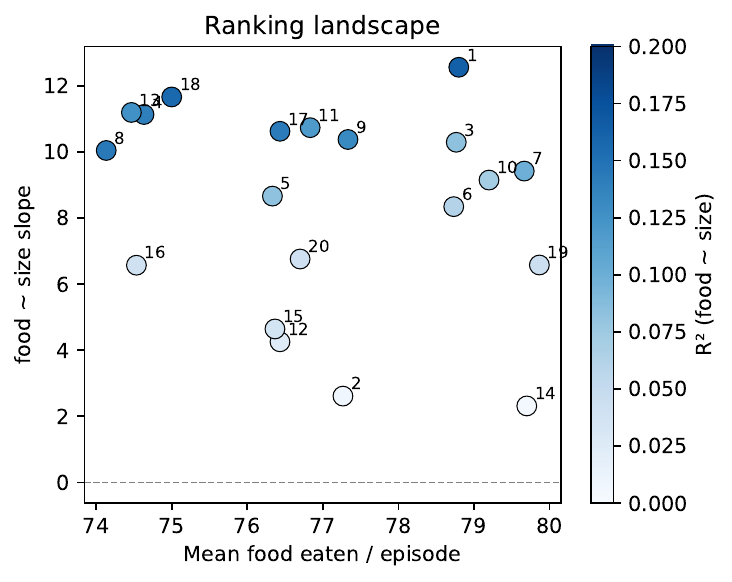}
\panel[0.49\textwidth]{C}{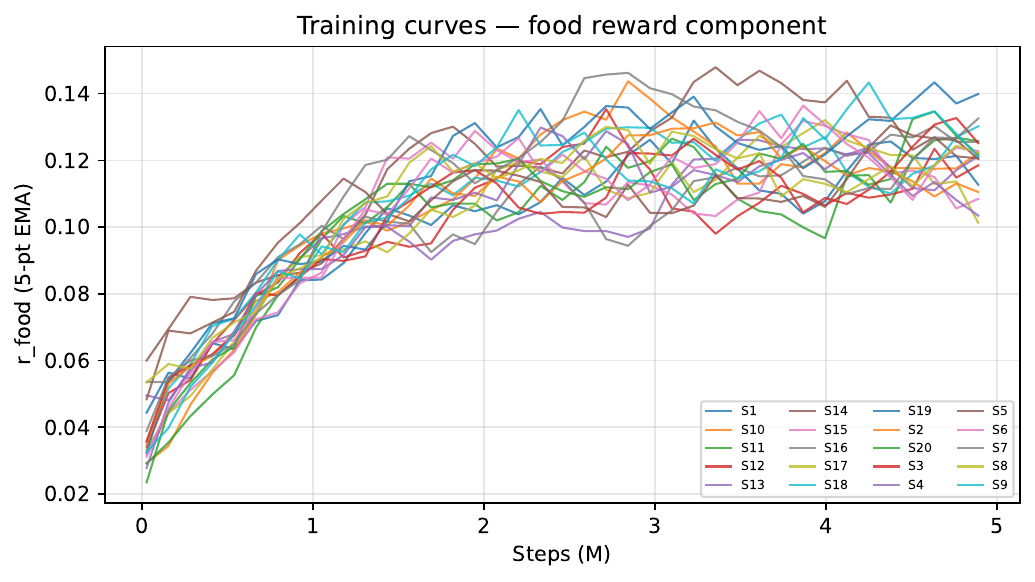}
\panel[0.49\textwidth]{D}{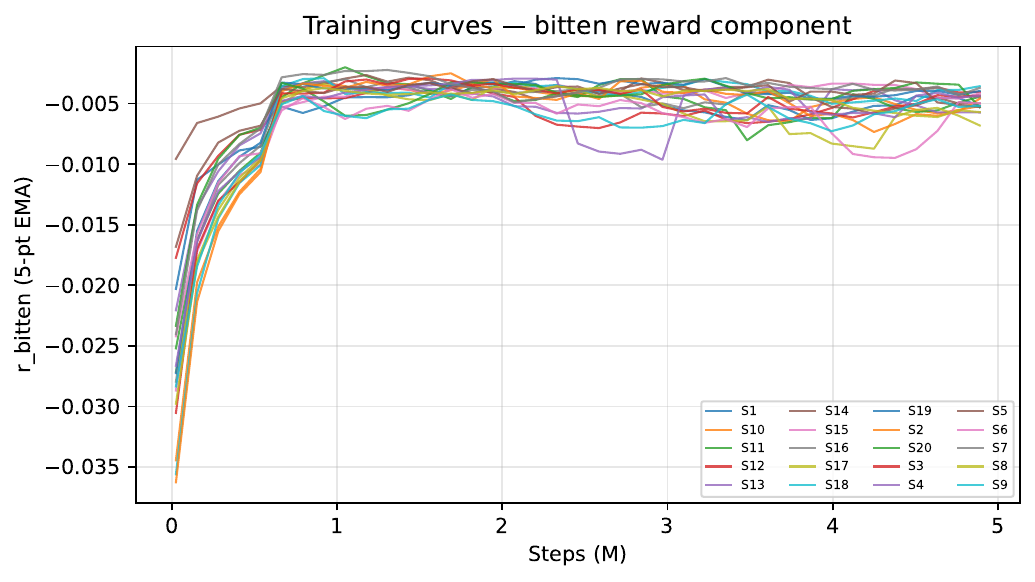}
\caption[]{
\textbf{Inter-seed reproducibility across 20 independent training runs (S1--S20, 5M steps each).}
\textbf{(A)} Total episode reward training curves (5-pt EMA) for all 20 seeds.
All seeds converge reliably to a similar performance band ($\sim$100--160) by 5M steps with no catastrophic failures.
\textbf{(B)} Seed ranking landscape: each point is one seed, positioned by mean food eaten per episode (x-axis, foraging efficiency) and the slope of a food$\sim$size linear regression (y-axis, strength of size-based dominance). Color encodes $R^2$ of food$\sim$size.
Seeds in the upper-right quadrant (high efficiency, strong dominance structure) were preferred for downstream analysis; S1 was selected as the canonical seed.
\textbf{(C)} Food reward component ($r_{\text{food}}$, 5-pt EMA) across all seeds, showing convergence to a fairly narrow band of foraging performance.
\textbf{(D)} Bitten reward component ($r_{\text{bitten}}$, 5-pt EMA) across all seeds.
The penalty is rapidly minimized from strongly negative values at initialization to near zero, indicating agents learn to avoid aggression costs over the course of training.
}
\label{fig:supp_seeds}
\end{figure*}

\begin{figure*}[!tbp]
\centering
\panelvstack[\PanelW][1mm]{A}{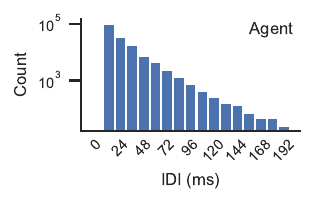}{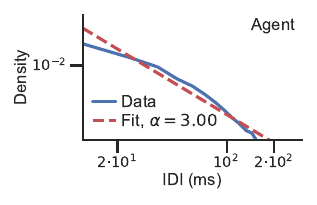}\hfill
\panel[\PanelW]{B}{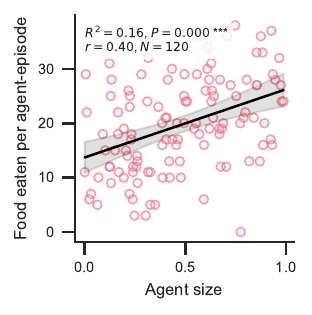}\hfill
\panel[\PanelW]{C}{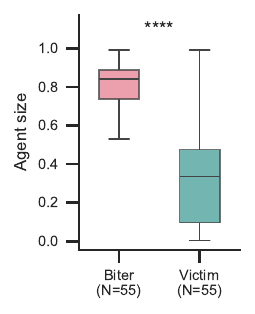}\hfill
\panel[\PanelW]{D}{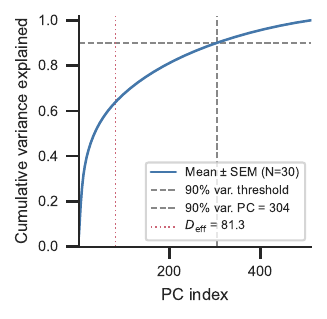}

\vspace{4pt}
\panelvstack[\PanelW][1mm]{E}{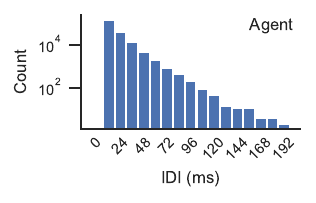}{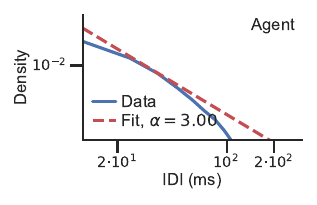}\hfill
\panel[\PanelW]{F}{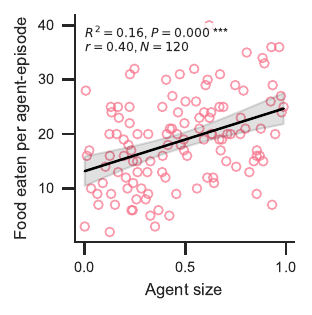}\hfill
\panel[\PanelW]{G}{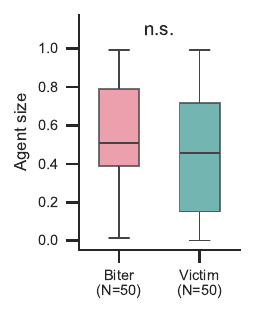}\hfill
\panel[\PanelW]{H}{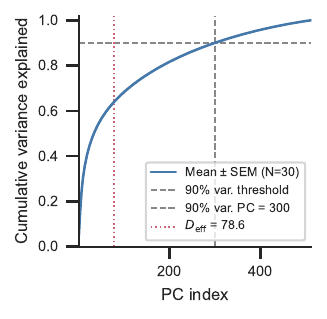}

\vspace{4pt}
\panelvstack[\PanelW][1mm]{I}{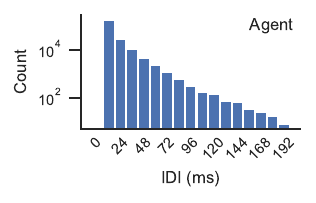}{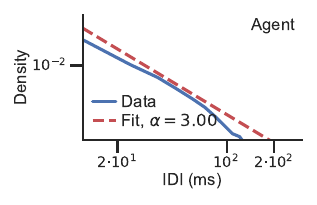}\hfill
\panel[\PanelW]{J}{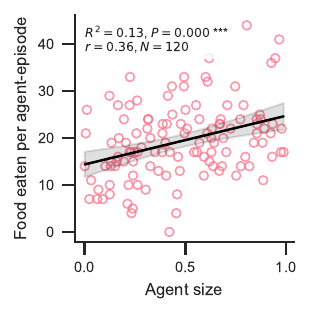}\hfill
\panel[\PanelW]{K}{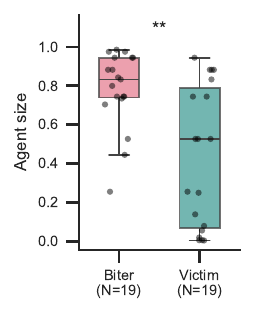}\hfill
\panel[\PanelW]{L}{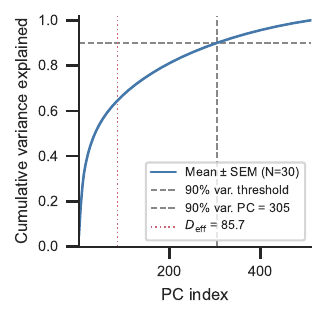}

\caption[]{
\textbf{Key result panels across the top-3 seeds (Rank 1: S1, Rank 2: S18, Rank 3: S9; rows A--D, E--H, I--L respectively).}
\textbf{(A,E,I)} Inter-discharge interval histogram and power-law fit (2-fish uniform-wide eval).
\textbf{(B,F,J)} Food eaten vs.\ agent size (4-fish patchy eval; Fig.~3M).
\textbf{(C,G,K)} Biter vs.\ bitten size by role boxplot (2-fish uniform-wide eval; Fig.~5M).
\textbf{(D,H,L)} RNN cumulative variance explained per episode, showing effective dimensionality (4-fish patchy eval; Fig.~6A).
Note that Rank 1 was used in the main manuscript.
}
\label{fig:supp_inter_seed}
\end{figure*}

See Fig.~\ref{fig:supp_interactions} for a schematic that summarizes the various types of dyadic agent interactions.

\subsubsection{PLSC details}
\label{supp:plsc_math}
Given two mean-centered data matrices $X \in \mathbb{R}^{n \times p}$ and $Y \in \mathbb{R}^{n \times q}$ (here, RNN hidden states of two agents across $n$ timesteps), PLSC computes the cross-covariance matrix
\[ C_{XY} = X^\top Y \in \mathbb{R}^{p \times q}. \]
Singular value decomposition (SVD) of this matrix,
\[ C_{XY} = U \, \Sigma \, V^\top, \]
yields orthogonal weight vectors $U$ and $V$ for $X$ and $Y$, respectively.
Projecting $X$ and $Y$ onto these weight vectors produces paired latent variables (canonical variates), whose sample correlations quantify the strength of shared structure along each dimension (e.g.\ PLSC1, the first-component correlation).

\clearpage
\subsection{Supplementary Figures}

The following figures provide additional training diagnostics and inter-seed reproducibility evidence.

\clearpage

\clearpage

\clearpage
\subsection{Simulation and Calibration Parameters}
\begin{table}[htbp!]
    \centering
    \begin{tabular}{|l|l|l|l|}
        \hline
        \textbf{Parameter} & \textbf{Symbol} & \textbf{Unit} & \textbf{Value} \\
        \hline
        \multicolumn{4}{|l|}{\textbf{Simulation Configuration}} \\
        \hline
        Prey radius & $R_{\text{prey}}$ & m & 0.0025 \\
        Agent radius & $R_{\text{agent}}$ & m & 0.01 \\
        Conductor contrast factor & $\chi$ & -- & -0.5 \\
        Default training arena range & -- & m & 0.40--3.00 \\
        One-patch arena size & -- & m & 1.50 $\times$ 1.50 \\
        Homing training arena range & -- & m & 0.75--1.50 \\
        Simulation rate & -- & Hz & 83 \\
        Electric wall-image scale & -- & -- & 0.95 \\
        \hline
        \multicolumn{4}{|l|}{\textbf{Sensor Thresholds}} \\
        \hline
        Ampullary min. threshold & $E_{\text{A,min}}$ & V/m & $2.0 \times 10^{-10}$ \\
        Ampullary max. threshold & $E_{\text{A,max}}$ & V/m & $2.0 \times 10^{-8}$ \\
        Mormyromast min. threshold & $E_{\text{M,min}}$ & V/m & $5.0 \times 10^{-8}$ \\
        Mormyromast max. threshold & $E_{\text{M,max}}$ & V/m & $5.0 \times 10^{-2}$ \\
        Knollenorgan min. threshold & $E_{\text{K,min}}$ & V/m & $2.0 \times 10^{-7}$ \\
        Knollenorgan log-saturation scale & $E_{\text{K,log max}}$ & V/m & $2.0 \times 10^{-4}$ \\
        \hline
        \multicolumn{4}{|l|}{\textbf{Active (Self-Image) Detection Ranges}} \\
        \hline
        Mormyromast prey detection range & $x_{\text{M,prey}}$ & m & 0.05 \\
        Mormyromast agent detection range & $x_{\text{M,agent}}$ & m & 0.10 \\
        \hline
        %
        %
        \multicolumn{4}{|l|}{\textbf{Passive Intrinsic Detection Ranges}} \\
        \hline
        Ampullary prey detection range & $x_{\text{A,prey}}$ & m & 0.04 \\
        Ampullary agent detection range & $x_{\text{A,agent}}$ & m & 0.08 \\
        \hline
        \multicolumn{4}{|l|}{\textbf{Passive Cons-EOD Detection Ranges}} \\
        \hline
        Knollenorgan agent detection range & $x_{\text{K,agent,consEOD}}$ & m & 1.00 \\
        \hline
        \multicolumn{4}{|l|}{\textbf{Calculated/Estimated Quantities}} \\
        \hline
        Agent EOD monopole source & $q_{\text{EOD}}$ & C & $\pm 1.11 \times 10^{-15}$ \\
        Agent EOD source separation & -- & cm & 1.0 \\
        Prey intrinsic dipole moment & $\mathbf{p}_{\text{prey}}$ & C$\cdot$m & $(0, 1.11 \times 10^{-24})$ \\
        Agent intrinsic dipole moment & $\mathbf{p}_{\text{agent}}$ & C$\cdot$m & $(1.11 \times 10^{-23}, 0)$ \\
        Cons-image scaling factor & $k_{\text{cons}}$ & - & $100$ \\
        \hline
    \end{tabular}
    \caption{Physics simulation and calibrated parameters used for the dynamic-sensing model. Distances are reported in meters unless otherwise specified. Electric-field thresholds are reported in V/m after unit conversion.}
    \label{tab:calibrated}
\end{table}

\subsection{Summary of evaluation assays}

\begin{table}[ht]
\centering
\footnotesize
\begin{tabular}{p{0.20\linewidth}p{0.40\linewidth}p{0.40\linewidth}}
\toprule
\textbf{Figure panels} & \textbf{Assay} & \textbf{Notes} \\
\midrule
Fig.~2A--D & Homing to a fixed emitting target in a 70 cm $\times$ 70 cm arena & 100 precomputed trials; compared with a non-electric ray baseline \\
Fig.~3A,B,K--P & Four-agent patchy foraging in a 70 cm $\times$ 70 cm arena & 30 episodes; heterogeneous agent sizes \\
Fig.~3C--J & Two-agent wide foraging in a 160 cm $\times$ 40 cm arena & 100 episodes; IDI and distance-dependent EOD analyses \\
Fig.~4A--F & Four-agent sensor-ablation foraging in patchy arenas & Single-sensor removals and reduced-sensor controls \\
Fig.~4G,H & Collective-sensing manipulations in a 40 cm $\times$ 40 cm arena & Self-EOD and conspecific-EOD components separated in the Mormyromast channel \\
Fig.~4I,J & EOD muting manipulations & EOD emission silenced in selected agents; food consumed and nearest-neighbor distance compared against baseline \\
Fig.~4K,L & Patchy versus uniform food distributions & Same trained policy evaluated under altered food layout \\
Fig.~4M,N & Iso-food patch-number sweep & Patch count 1--5; iso series holds total food constant; free series holds food-per-patch constant \\
Fig.~5A--H & Two-agent one-patch competition assay & 100 trials per trained-agent size condition \\
Fig.~5I,J & Patch-confined bot control & 300-condition grid crossing size, EOD rate, movement state, and initial position \\
Fig.~5K--P & Two-agent square interaction assay & 70 cm $\times$ 70 cm arena; pairwise approach, chasing, and biting analyses \\
Fig.~6A & Four-agent recurrent-state PCA & Hidden states from patchy foraging episodes \\
Fig.~6B,C & Group-size recurrent dimensionality assay & Same policy evaluated with different numbers of agents \\
Fig.~6D & Homing recurrent-state decoding & Knollenorgan-defined and straight-line target angles decoded from hidden states \\
Fig.~6E,F & Two-agent recurrent-state decoding & Split by sensory range and pooled across timesteps \\
Fig.~6G,H & Two-agent shared-state PLSC & In-range versus out-of-range pairwise comparisons \\
Fig.~6I,J & Two-agent recurrent-state power analysis & Episode-level wideband power associated with social proximity and relative size \\
\bottomrule
\end{tabular}
\caption{Condensed summary of the evaluation assays used for each main figure.}
\label{tab:evaluation_assays}
\end{table}

\vspace{1em}

\clearpage
\subsection{List of recurrent-state decoding targets}
\begin{table}[ht]
\centering
\footnotesize
\begin{tabular}{p{0.26\linewidth}p{0.13\linewidth}p{0.12\linewidth}p{0.40\linewidth}}
\toprule
\textbf{Target} & \textbf{Symbol} & \textbf{Type} & \textbf{Description} \\
\midrule
X Position & $x$ & Scalar & Allocentric x position \\
Y Position & $y$ & Scalar & Allocentric y position \\
Distance to Nearest Agent & $d_{\mathrm{agent}}$ & Scalar & Distance to closest conspecific \\
Distance to Closest Food & $d_{\mathrm{food}}$ & Scalar & Distance to nearest food item \\
Angle to Closest Agent & $\alpha_{\mathrm{agent}}$ & Circular & Bearing to closest conspecific \\
Angle to Closest Food & $\alpha_{\mathrm{food}}$ & Circular & Bearing to nearest food item \\
Actual Turn & $\omega$ & Scalar & Realized angular change per timestep \\
Agent Size & $s_{\mathrm{agent}}$ & Scalar & Scalar size context of the focal agent \\
Energy & $e$ & Scalar & Internal energy state, when present in the rollout table \\
Agents in Knollen Range & $n^K_{\mathrm{agent}}$ & Count & Number of conspecifics within Knollenorgan range \\
Size of Nearest Agent & $s_{\mathrm{agent,1}}$ & Scalar & Size metadata for the nearest conspecific \\
Distance to Second Nearest Agent & $d_{\mathrm{agent,2}}$ & Scalar & Distance to the second-nearest conspecific \\
Size of 2nd Nearest Agent & $s_{\mathrm{agent,2}}$ & Scalar & Size metadata for the second-nearest conspecific \\
Food Count Centered (5cm) & $n^M_{\mathrm{food}}$ & Count & Number of food items within the centered 5 cm local window \\
Food Count Front (5cm) & $n_{\mathrm{food,front}}$ & Count & Number of food items in the 5 cm front sector \\
Food Count Behind (5cm) & $n_{\mathrm{food,back}}$ & Count & Number of food items in the 5 cm rear sector \\
Food Count Left (5cm) & $n_{\mathrm{food,left}}$ & Count & Number of food items in the 5 cm left sector \\
Food Count Right (5cm) & $n_{\mathrm{food,right}}$ & Count & Number of food items in the 5 cm right sector \\
Agent Nearby & $I^M_{\mathrm{agent}}$ & Boolean & Whether a conspecific is within the nearby-agent threshold \\
Was Bitten & $I_{\mathrm{bitten}}$ & Boolean & Whether the agent was bitten on that timestep \\
Agent in Knollen Range & $I^K_{\mathrm{agent}}$ & Boolean & Whether any conspecific is within Knollenorgan range \\
Mormyromast Field Magnitude & $\log|\vec{E}_{\mathrm{M}}|$ & Scalar & Log-transformed Mormyromast field magnitude, when center-field columns are available \\
Ampullary Field Magnitude & $\log|\vec{E}_{\mathrm{A}}|$ & Scalar & Log-transformed Ampullary field magnitude, when center-field columns are available \\
Knollen Error Angle & $\alpha^K_{\mathrm{agent}}$ & Circular & Orientation error relative to the Knollenorgan field direction \\
Distance to Wall & $d_{\mathrm{wall}}$ & Scalar & Distance from the focal agent to the nearest arena wall \\
Mormyromast Error Angle & $\alpha^M_{\mathrm{agent}}$ & Circular & Homing-task error angle for the Mormyromast field \\
Ampullary Error Angle & $\alpha^A_{\mathrm{agent}}$ & Circular & Homing-task error angle for the Ampullary field \\
Knollen Error Angle to Nearest Agent & $\theta^K_{\mathrm{error}}$ & Circular & Homing-task error angle for the nearest-agent Knollen field \\
\bottomrule
\end{tabular}
\caption{Recurrent-state decoding targets used in the analyses. Targets absent from a given assay were omitted from that assay-specific decoding fit.}
\label{tab:features}
\end{table}

\newpage

\newpage
\subsection{Sensor Calibration Calculations}
\label{supp:calibration_calculations}
\subsubsection{Dipole Moment for a Given Maximum Field at Distance \( x \)}

The electric field ($\bm{E}$) of a dipole at a point $\bm{r}$ (where $x = |\bm{r}|$) is given by the general vector formula:
\[
\bm{E}(\bm{r}) = \frac{1}{4 \pi \varepsilon_0} \left[ \frac{3(\bm{p} \cdot \hat{\bm{r}})\hat{\bm{r}} - \bm{p}}{x^3} \right]
\]
where $\bm{p}$ is the dipole moment vector and $\hat{\bm{r}}$ is the unit vector in the direction of $\bm{r}$.
Alternatively, in spherical coordinates the magnitude of the electric field at a point ($x, \theta$) is:
\[
E(x, \theta) = \frac{p}{4 \pi \varepsilon_0 x^3} \sqrt{3\cos^2\theta + 1}
\]
where $\theta$ is the angle between the dipole moment vector $\bm{p}$ and the position vector $\bm{r}$.

The magnitude of the electric field ($E$) is maximized when $\cos\theta = \pm 1$, which corresponds to an angle of $\theta = 0^{\circ}$ or $\theta = 180^{\circ}$, i.e. directly along the axis of the dipole. 
The field magnitude along the axis is then:
\[
E_{\text{max}} = \frac{p}{2 \pi \varepsilon_0 x^3}
\]

Therefore, the dipole moment \( p \) required to generate a maximum electric field \( E_{\text{max}} \) at a distance \( x \) is:
\[
p = 2 \pi \varepsilon_0 x^3 E_{\text{max}}
\]
where $p$ is the dipole moment (C$\cdot$m), $x$ is the distance from the dipole where the field is maximized (m), $E_{\text{max}}$ is the required electric field magnitude (V/m), and $\varepsilon_0$ is the permittivity of free space ($8.854 \times 10^{-12}$ F/m).

\newpage
\subsubsection{Dipole Moment Required to Induce a Specific Field at Its Own Location}

Active sensing involves three steps: a source dipole ($\bm{p}$) creates a field, this field induces a new dipole ($\bm{p}'$) on a conductor (other agent or prey), and this induced dipole in turn creates a field that the source's sensors detect.

To calibrate the strength of the EOD source, we find the source dipole moment ($p$) such that the maximum induced field at the source's location, $E_{\text{induced}}$, is equal to a desired value, $E_{\text{max}}$.

The electric field at the conductor due to the source dipole is given by the on-axis formula:
\[
E_{\text{source}} = \frac{p}{2 \pi \varepsilon_0 x^3}
\]

This source field induces a dipole moment ($\bm{p}'$) on the conductor, which is assumed to be a sphere of radius $R$:
\[
p' = 3 \varepsilon_0 V \cdot \chi \cdot E_{\text{source}}
\]
where $V = \frac{4}{3} \pi R^3$ is the volume of the conductor. 

Substituting the expression for $E_{\text{source}}$ and simplifying yields:
\[
p' = 3 \varepsilon_0 \left( \frac{4}{3} \pi R^3 \right) \cdot \chi \cdot \frac{p}{2 \pi \varepsilon_0 x^3} = \frac{2 R^3 \cdot \chi \cdot p}{x^3}
\]

The induced dipole ($\bm{p}'$) then creates its own on-axis electric field at the location of the original source dipole. The magnitude of this induced field is:
\[
E_{\text{induced}} = \frac{p'}{2 \pi \varepsilon_0 x^3}
\]
Substituting the simplified expression for $p'$ into this equation, we get:
\[
E_{\text{induced}} = \frac{1}{2 \pi \varepsilon_0 x^3} \left( \frac{2 R^3 \cdot \chi \cdot p}{x^3} \right) = \frac{R^3 \cdot \chi \cdot p}{\pi \varepsilon_0 x^6}
\]

To calibrate the source EOD dipole strength, we set this induced field equal to the sensor's minimum detection threshold, $E_{\text{max}}$ and solve for the required source dipole moment, $p$:
\[
p = \frac{E_{\text{max}} \pi \varepsilon_0 x^6}{R^3 \cdot \chi}
\]
where,
$p$ is the required dipole moment of the source (C$\cdot$m),
$E_{\text{max}}$ is the target electric field magnitude at the source dipole (V/m),
$x$ is the separation between the source dipole and the conductor (m),
$R$ is the conductor’s radius (m),
$\chi$ is the conductor contrast factor,
$\varepsilon_0$ is the permittivity of free space ($8.854 \times 10^{-12}$ F/m).

\newpage
\subsubsection{Self-EOD vs. Induced-dipole field strength at a Mormyromast sensor (on-axis, collinear)}
\label{supp:range_checks}

Consider a source dipole at $(0,0)$ with moment $\bm p=p\,\hat{\bm y}$, a spherical object of radius $R$ centered at $(0,x_{\text{obj}})$, and a Mormyromast-type sensor on the source body at $(0,r)$ with outward normal $\hat{\bm y}$. 
All positions are collinear along $\hat{\bm y}$ and fields are evaluated on-axis.

\paragraph{Self field at the sensor:}
On-axis self-EOD field magnitude at $(0,r)$:
\[
E_{\text{self}}=\left|\frac{1}{2\pi\varepsilon_0}\,\frac{p}{r^3}\right|.
\]

\paragraph{Induced field at the sensor:}
Field at the object’s center:
\[
E_{\text{source}\to\text{obj}}=\left|\frac{1}{2\pi\varepsilon_0}\,\frac{p}{x_{\text{obj}}^3}\right|.
\]
Assuming object is a sphere with contrast factor $\chi$ the induced dipole moment is
\[
p' \;=\; 4\pi\varepsilon_0 R^3\,\chi\,E_{\text{source}\to\text{obj}}
\;=\; \frac{2 R^3 \chi}{x_{\text{obj}}^3}\,p.
\]

The field magnitude at $(0,r)$ due to the induced dipole at $(0,x_{\text{obj}})$ is
\[
E_{\text{ind}}=\left|\frac{1}{2\pi\varepsilon_0}\,\frac{p'}{|x_{\text{obj}}-r|^3}\right|
=\frac{1}{2\pi\varepsilon_0}\,\frac{1}{|x_{\text{obj}}-r|^3}\left(\frac{2 R^3 \chi}{x_{\text{obj}}^3}p\right).
\]

\paragraph{Ratio (self vs.\ induced).}
Taking the ratio of magnitudes,
\[
\frac{E_{\text{self}}}{E_{\text{ind}}}
\;=\;
\frac{\displaystyle \frac{p}{2\pi\varepsilon_0 r^3}}
{\displaystyle \frac{1}{2\pi\varepsilon_0}\frac{2 R^3 \chi}{x_{\text{obj}}^3}\frac{p}{|x_{\text{obj}}-r|^3}}
\;=\;
\boxed{\;\frac{x_{\text{obj}}^{3}\,|x_{\text{obj}}-r|^{3}}{2\,\chi\,R^{3}\,r^{3}}\,.\;}
\]
In the common regime $r\ll x_{\text{obj}}$, $|x_{\text{obj}}-r|\approx x_{\text{obj}}$ and
\[
\frac{E_{\text{self}}}{E_{\text{ind}}}\;\approx\;\frac{x_{\text{obj}}^{6}}{2\,\chi\,R^{3}\,r^{3}}.
\]

\noindent For a prey object that is $x_{obj}=5$ cm away, with $R=0.25$ cm radius, \\ $|\chi| = 0.5$, and agent radius $r=1$ cm,
$\frac{E_{\text{self}}}{E_{\text{ind}}}\;\approx\; 10^6$
\\

\noindent For a post-baseline-subtraction sensing dynamic range of 10 (``one order of magnitude'') on $E_{ind}$, i.e.:
$E^B_{ind} / E^A_{ind} = 10$, implies a distance range of $x^A_{obj} / x^B_{obj} = 10^{1/6} = 1.47 = 1/(0.68)$

\newpage
\subsubsection{Self-EOD vs. Cons-EOD field strength at a Mormyromast sensor (on-axis, collinear)}

Given two agents, one at $(0,0)$ and the conspecific agent at $(0, x_{cons})$, we will calculate the ratio between the field due to self-EOD $E_{\text{self-EOD}}$ and due to cons-EOD $E_{\text{cons-EOD}}$ at a Mormyromast-type sensor located at $(0, r)$ with a radially-outward pointing normal. 

For simplicity, we assume that a point dipole $\bm p = p\,\hat{\bm y}$ at the first agent's center generates the \emph{self-EOD}. 
When a conspecific emits (receiver is not emitting), the
conspecific dipole (assumed collinear and aligned) at $y=x_{\text{cons}}>r$ on the $+\hat{\bm y}$
axis. 

The on-axis dipole field is given by
\[
E_y^{(\text{dip})}(\Delta y)=\frac{1}{2\pi\varepsilon_0}\;p\;\frac{\operatorname{sgn}(\Delta y)}{|\Delta y|^3},
\]

$E_{\text{self-EOD}}$, the magnitude of the self-EOD field at the sensor is given by:

\[
E_{\text{self-EOD}}
=\left|\frac{1}{2\pi\varepsilon_0}\,\frac{p_{\text{self}}}{r^3}\right|.
\]

$E_{\text{cons-EOD}}$, the magnitude of the cons-EOD field, due to a dipole at $(0, x_{\text{cons}})$, at the same sensor is given by:
\[
E_{\text{cons-EOD}}
=\left|\frac{1}{2\pi\varepsilon_0}\,\frac{p_{\text{cons}}}{|x_{\text{cons}}-r|^3}\right|.
\]

Assuming same dipole strength, their ratio is given by
$$
\frac{ E_{\text{self-EOD}} }{ E_{\text{cons-EOD}} } = 
\frac{|x_{\text{cons}}-r|^3}{ r^3 }
$$

\noindent For agents separated by $ x_{cons} \approx 10$ cm and a agent body radius $r \approx 1$ cm, \\
$ E_{\text{self-EOD}} / E_{\text{cons-EOD}} = 9^3 = 729$.

\end{document}